\documentclass[journal]{IEEEtran}
\usepackage{multirow}
\usepackage{booktabs}
\usepackage{bbm}
\usepackage{algorithmic}
\usepackage{algorithm}
\usepackage{amsthm,amsmath,amssymb}
\usepackage{mathrsfs}
\usepackage{graphicx}
\usepackage{amsfonts}
\usepackage{bm}
\usepackage{cite}
\usepackage{amssymb}
\usepackage{doi}
\usepackage{color}
\usepackage{url}
\usepackage{hyperref}
\usepackage{makecell}
\usepackage{stackengine}
\usepackage[final]{changes}
\newcommand\xrowht[2][0]{\addstackgap[.5\dimexpr#2\relax]{\vphantom{#1}}}
\ifCLASSOPTIONcompsoc
\usepackage[caption=false,font=normalsize,labelfont=sf,textfont=sf]{subfig}
\else
\usepackage[caption=false,font=footnotesize]{subfig}
\fi

\definecolor{ccr}{RGB}{10,110,150}
\hypersetup{hypertex=true,
    colorlinks=true,
    linkcolor=ccr,
    anchorcolor=ccr,
    citecolor=ccr}

\ifCLASSINFOpdf
\else
\fi
\begin{document}
\title{Source-free Semantic Regularization Learning for Semi-supervised Domain Adaptation}
\author{Xinyang~Huang,~Chuang~Zhu\IEEEauthorrefmark{2}\thanks{\dag~Chuang Zhu is the corresponding author.},~\IEEEmembership{Member,~IEEE,}~Ruiying~Ren,~Shengjie~Liu,
and~Tiejun~Huang,~\IEEEmembership{Senior~Member,~IEEE}
\thanks{Supported by the National Key R\&D Program of China (2021ZD0109802), the National Natural Science Foundation of China (81972248), the High-performance Computing Platform of BUPT, and BUPT Innovation Program (2024-YC-T031).}
\thanks{X. Huang, C. Zhu, R. Ren, and S. Liu are with the School of Artificial Intelligence, Beijing University of Posts and Telecommunications, Beijing, China.
E-mail: hsinyanghuang7@gmail.com}
\thanks{T. Huang is with the BAAI and Institute for Artificial Intelligence, Peking University, Beijing, China.
Huang is also with the National Engineering Research Center of Visual Technology, School of Computer Science, Peking University.}}
\maketitle

\begin{abstract}
Semi-supervised domain adaptation (SSDA) has been extensively researched due to its ability to improve classification performance and generalization ability of models by using a small amount of labeled data on the target domain. 
However, existing methods cannot effectively adapt to the target domain due to difficulty in fully learning rich and complex target semantic information and relationships.
In this paper, we propose a novel SSDA learning framework called semantic regularization learning (SERL), which captures the target semantic information from multiple perspectives of regularization learning to achieve adaptive fine-tuning of the source pre-trained model on the target domain.
SERL includes three robust semantic regularization techniques.
Firstly, semantic probability contrastive regularization (SPCR) helps the model learn more discriminative feature representations from a probabilistic perspective, using semantic information on the target domain to understand the similarities and differences between samples.
Additionally, adaptive weights in SPCR can help the model learn the semantic distribution correctly through the probabilities of different samples.
To further comprehensively understand the target semantic distribution, we introduce hard-sample mixup regularization (HMR), which uses easy samples as guidance to mine the latent target knowledge contained in hard samples, thereby learning more complete and complex target semantic knowledge.
Finally, target prediction regularization (TPR) regularizes the target predictions of the model by maximizing the correlation between the current prediction and the past learned objective, thereby mitigating the misleading of semantic information caused by erroneous pseudo-labels.
Extensive experiments on three benchmark datasets demonstrate that our SERL method achieves state-of-the-art performance.
\end{abstract}

\begin{IEEEkeywords}
Domain adaptation, Semi-supervised learning, Source-free, Semantic regularization.
\end{IEEEkeywords}

\IEEEpeerreviewmaketitle

\section{Introduction}
\label{introduction}
\IEEEPARstart{I}{n} recent years, deep neural networks (DNN) have brought a series of breakthroughs in many computer vision tasks, such as image classification \cite{krizhevsky2012imagenet, he2016deep, rastegari2016xnor, krizhevsky2017imagenet, qiao2019transductive,yan2016image,wang2016csps}, semantic segmentation \cite{long2015fully,gao2022fbsnet,yan2019semantic,minaee2021image,li2023muva, wang2023seggpt}.
However, to achieve satisfactory results, the large number of sample labels required for deep neural network training is costly and time-consuming.
Therefore, domain adaptation (DA) \cite{pan2010domain,patel2015visual,mei2020instance,zhang2021self} is proposed by generalizing the knowledge learned from the source domain with rich labels to the target domain with no or few labels.
Domain adaptation can be simply divided into unsupervised domain adaptation (UDA) \cite{ganin2015unsupervised,deng2021informative,wang2022cross,jing2022adversarial,lu2021discriminative, zhao2020review,zuo2023dual,ding2023unsupervised} and semi-supervised domain adaptation (SSDA) \cite{saito2019semi,li2021cross,li2021ecacl,qin2022semi,xu2022semi,yan2022multi, yu2023semi,huang2023semi,li2023adaptive,li2024inter,chen2023semi} according to access to target labels during training. 
This paper focuses on SSDA, which performs significantly better than UDA when given a small number of labeled target samples.
It can utilize a small number of labels on the target domain to expand semantic information and learn semantic knowledge of target samples of the same category to achieve domain alignment.

\begin{figure}[t]
	\centering
\includegraphics[width=0.85\linewidth]{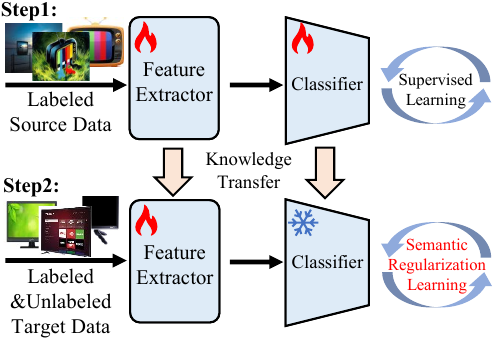}
	\caption{The learning scenario of our SERL framework.
Different from the training paradigm of most existing SSDA methods, we adopt a source-free training strategy.
The source model comprises a feature extractor and a classifier initialized on the source domain.
We focus on improving the target domain adaptation stage of the model.
In the target domain adaptation stage, SERL freezes the classifier module and fine-tunes the feature extractor module through semantic regularization learning.}
	\label{intro1}
    \vspace{-1.5em}
\end{figure}

Due to its advantages of practical significance, SSDA has attracted increasing attention and has been widely studied.
However, SSDA also has its specific challenges and issues.
First, the training of the supervised model only uses a small number of labeled target samples. 
The model can only learn the extremely limited target domain knowledge and cannot generate a highly discriminative knowledge representation for the target domain \cite{saito2019semi,li2023adaptive}.
At the same time, due to many labeled source samples, the feature representation learned by the model is biased toward the source domain \cite{kim2020attract}.
To address these issues, existing methods \cite{saito2019semi,li2021cross,li2021ecacl,qin2022semi,xu2022semi,yan2022multi,yu2023semi,huang2023semi,li2023adaptive,li2024inter} have proposed their solutions to address these challenges and have witnessed significant performance improvements.
MCL \cite{yan2022multi} learns the consistency between samples, but it ignores the learning of target semantic information.
ProML \cite{huang2023semi} utilizes target labels by constructing prototypes, but the semantic information contained in them is very limited due to the scarcity of labeled target samples.
Due to the complexity of semantic information between target samples, the knowledge representation learned by existing methods still needs to be improved.
This complex semantic information $i.e.$, category knowledge representation on the target domain can better bridge the distribution differences between domains, encouraging the model to generate domain-invariant but differentiated target features when adapting.

\begin{figure}[t]
	\centering
 \includegraphics[width=0.95\linewidth]{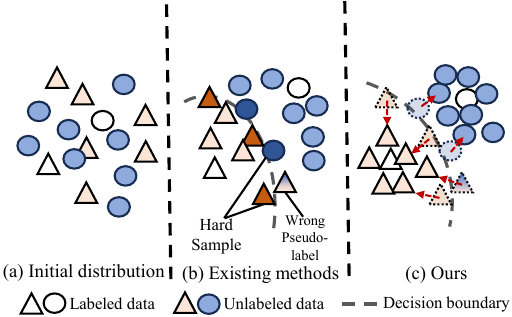}
	\caption{The motivation of our SERL.
(a) Due to the scarcity of target semantic labels during training, most existing SSDA methods have shortcomings in target semantic learning, resulting in models only learning limited knowledge (e.g., only the relationships between samples) on the target domain.
When more complex relationships exist on the target domain, such as hard and noisy samples, the model may perform poorly due to a lack of understanding of semantic information.
(b) Our SERL utilizes the semantic information learned on the target domain from the perspective of semantic regularization to constrain the feature representation of the model further, thereby adapting to more complex target domain distributions.}
	\label{intro2}
\end{figure}

In this paper, we present a novel SSDA learning framework, named semantic regularization learning (SERL), which is proposed to tackle the challenges of the SSDA tasks.
As shown in Figure \ref{intro1}, different from the training paradigm of most existing SSDA methods, this paper considers a source-free scenario, $i.e.$, in which target domain adaptation is performed using the source domain pre-trained model \cite{liang2020we}.
Unlike UDA, SSDA can obtain a small amount of labeled data on the target domain, so it can better adapt to this source-free scenario.
SERL provides regularization constraints from different perspectives by fully learning the target semantic information, which can enrich the understanding of the accurate distribution of the target domain and thus better learn the knowledge of the target domain, as shown in Figure \ref{intro2}.

Specifically, we propose semantic probability contrastive regularization (SPCR), which helps the model aggregate features of similar samples according to the distribution of target semantic information and keep features of heterogeneous samples away from each other.
This method forces the model to learn more discriminative semantic knowledge on the target domain from the probability perspective.
At the same time, SPCR uses adaptive weights to assign lower weights to low-confidence samples by combining the confidence of contrasting examples to reduce the impact of erroneous semantic information and help the model learn the correct target distribution.
Furthermore, hard samples are crucial to fully understand the target semantic distribution \cite{xuan2020hard,zuo2021challenging,liu2022complementary,xiong2023confidence}.
However, existing SSDA methods ignore exploring hard samples due to their complex knowledge distribution.
To fill this gap, we further explore the complex target relationships of hard target samples through hard-sample mixup regularization (HMR).
After screening out easy and hard samples using the classifier prototype, we use easy sample guidance to learn these hard samples.
Specifically, HMR uses the regularization constraint of mixup \cite{zhang2017mixup} to mix easy samples with hard samples.
This method further explores the potential knowledge of hard samples through the guidance of easy samples and further helps the model learn more complex target semantic information.
Finally, even if we consider the discriminative knowledge representation and hard sample information of the target domain, there will still be bias in semantic learning when there is much noise in the target pseudo-labels.
To reduce this misleading semantic information caused by noisy pseudo-labels, we minimize the impact of noisy pseudo-labels from the perspective of target prediction regularization (TPR).
Inspired by \cite{liu2020early, yi2023source}, we use the early prediction of samples to constrain the probability output of the model during the adaptation stage to encourage the model to follow early target sample predictions and alleviate overfitting of erroneous semantic information on the target domain.

In summary, our main contributions are as follows:
\begin{itemize}
	\item We propose a novel SSDA framework called semantic regularization learning (SERL).
The proposed SERL considers fully utilizing and learning semantic knowledge on the target domain to achieve cross-domain adaptation when fine-tuning the source model on the target domain.
	\item To fully utilize the semantic relationships of the target domain, we propose three regularization methods, $i.e.$, semantic probability contrastive regularization, hard-sample mixup regularization, and target prediction regularization, to constrain the performance of the model on the target domain through semantic regularization strategies and further learn the knowledge of the target domain.
	\item Extensive experiments conducted on three standard benchmark datasets, including DomainNet \cite{pei2018multi}, Office-Home \cite{venkateswara2017deep}, and Office-31 \cite{saenko2010adapting}, have shown that our method has significant advantages over previous state-of-the-art SSDA methods.
\end{itemize}

The paper is structured as follows:
In Section \ref{related work}, we provide an overview of prior research related to our work.
Section \ref{methodology} introduces and describes the proposed algorithm for semi-supervised domain adaptation.
In Section \ref{experiment}, we conduct comparative experiments to evaluate the performance of the proposed method.
Finally, the conclusions of our approach are presented in Section \ref{conclusion}.

\section{RELATED WORK}
\label{related work}
\subsection{Unsupervised Domain Adaptation}
To solve the problem that traditional supervised learning requires much manual annotation, unsupervised domain adaptation (UDA) aims to transfer knowledge from a fully labeled source domain to an unlabeled target domain.
In recent years, various methods have been proposed for UDA, and adequate progress has been achieved.
Commonly used methods mainly include maximum mean difference (MMD) \cite{gretton2012kernel}, whose basic idea is to achieve migration from the source domain to the target domain by minimizing the distance between feature distributions.
DANN \cite{ganin2016domain} and JAN \cite{long2017deep} further proposed using the MMD criterion to learn transfer networks by cross-region alignment of multiple region-specific layers.
CORAL \cite{sun2017correlation} and DUCDA \cite{zhuo2017deep} proposed minimizing the domain shift by aligning the second-order statistics of the source and target distributions.
Meanwhile, with the development of generative adversarial networks, many recent works \cite{ganin2015unsupervised,tzeng2017adversarial,zhang2021joint,xie2018learning,shen2018wasserstein,ge2023unsupervised,jing2022adversarial,shermin2020adversarial} have used adversarial learning for domain alignment so that knowledge from classifiers trained on labeled source samples can be effectively transferred to the target domain.
In addition, considering the perspective of conditional distributions, many related works \cite{chen2019progressive,pan2019transferrable,zhong2021does} have proven that learning conditional distributions is of good help in reducing the differences in the alignment of classification domains, thereby improving the adaptability between domains.
Although the UDA method has been successfully used in many practical applications, it takes work to accurately describe the conditional distribution of target features due to the significant differences between some source domains and target domains and the unreachability of target labels.
Therefore, the potential of the UDA method in practical applications is limited compared to the SSDA method.

\subsection{Semi-supervised Domain Adaptation}
Semi-supervised domain adaptation (SSDA) aims to utilize a small number of labeled samples on the target domain. Compared with UDA, the classification performance and generalization ability of the model on the target domain can be significantly improved due to the access to labeled target samples.
At present, SSDA has made much adequate progress, and the methods used in many works can be roughly divided into cross-domain alignment methods, adversarial training methods, and semi-supervised learning methods.
In cross-domain alignment, many related works \cite{li2021ecacl,yi2023source,singh2021improving,yang2021deep} integrate various complementary domain alignment technologies.
G-ABC \cite{li2023adaptive} further achieves semantic alignment by forcing the transfer from labeled source and target data to unlabeled target samples.
In addition, IDMNE \cite{li2024inter} is proposed to incorporate the label information of labeled samples into the model to learn cross-domain class feature alignment.
Utilizing the idea of adversarial training, many related methods \cite{saito2019semi,qin2021contradictory,jiang2020bidirectional,kim2020attract,li2021cross,qin2022semi} solve the SSDA problem by minimizing the entropy between the class prototype and adjacent unlabeled target domain samples to achieve the effect of adversarial training.
To solve SSDA through the idea of semi-supervised learning, MCL \cite{yan2022multi} and ProML \cite{huang2023semi} further help the model understand the target domain that lacks a large number of labels through consistency regularization.
Unlike most existing methods, DEEM \cite{ma2022context} considers a source-free \cite{liang2020we} scenario and proposes a self-distillation method to improve entropy minimization and help label propagation of unlabeled samples on the target domain.
However, the above existing methods all need to pay more attention to the importance of profoundly exploring the semantic information of the target domain.
This paper starts from the perspective of semantic regularization learning and proposes the SERL framework, which helps the model more comprehensively adapt to the actual target domain distribution by standardizing the knowledge representation learned by the model on the target domain.

\section{METHODOLOGY}
\label{methodology}
\subsection{Preliminaries and Overview}
In semi-supervised domain adaptation (SSDA), the model is expected to generalize well on the target domain with fully labeled source samples and a small number of labeled target samples.
Specifically, the source domain dataset ${\mathcal{S} = \{x^s_i, y^s_i\}^{N_s}_{i=1}}$ contains fully labeled data, ${\mathcal{L} = \{x^l_i, y^l_i\}^{N_l}_{i=1}}$ contains a small amount of labeled data of the target domain, where $N_s$ and $N_l$ are the source domain and target domain dataset size respectively.
Here, $x_i^s$ and $x_i^l$ represent the labeled source image and target image data, respectively, and \replaced{$y_i^s$ and $y_i^l$}{$y^s$ and $y^l$} represent the corresponding labels.
In addition to the labeled data, there is also an unlabeled target image set ${\mathcal{U} = \{x^u_i\}^{N_u}_{i=1}}$ for adaptation on the target domain, which contains unlabeled target image data, usually $N_u\gg N_l$.
The overall objective used to optimize the model can be expressed as a combination of the loss of the base model and the additional loss, as follows:
\begin{equation}
\label{allloss}
L_\mathrm{all}=L_\mathrm{base}+\lambda_\mathrm{prob}L_\mathrm{prob}+\lambda_\mathrm{mix}L_\mathrm{mix}+\lambda_\mathrm{pre}L_\mathrm{pre},
\end{equation}
where $\lambda_\mathrm{prob}$, $\lambda_\mathrm{mix}$ and $\lambda_\mathrm{pre}$ are scalar hyper-parameters of the loss weights and $L_\mathrm{prob}$, $L_\mathrm{mix}$, and $L_\mathrm{pre}$ represent semantic probability contrastive regularization, hard-sample mixup regularization and target prediction regularization respectively.

For the model trained on the source domain, we first use the cross-entropy loss to train the feature extractor $g(\cdot)$ and the linear classifier $f(\cdot)$.
For the source data ${\mathcal{S} = \{x^s_i, y^s_i\}^{N_s}_{i=1}}$, we employ
the standard cross-entropy objective:
\begin{equation}
\begin{aligned}
L_\mathrm{s}=\frac1{N_{s}}\sum_{i=1}^{N_{s}}L_{CE}(p(y_i^{s}|x_i^{s}),y_i^{s}).
\end{aligned}
\end{equation}

\begin{figure*}[t]
	\centering
\includegraphics[width=1\linewidth]{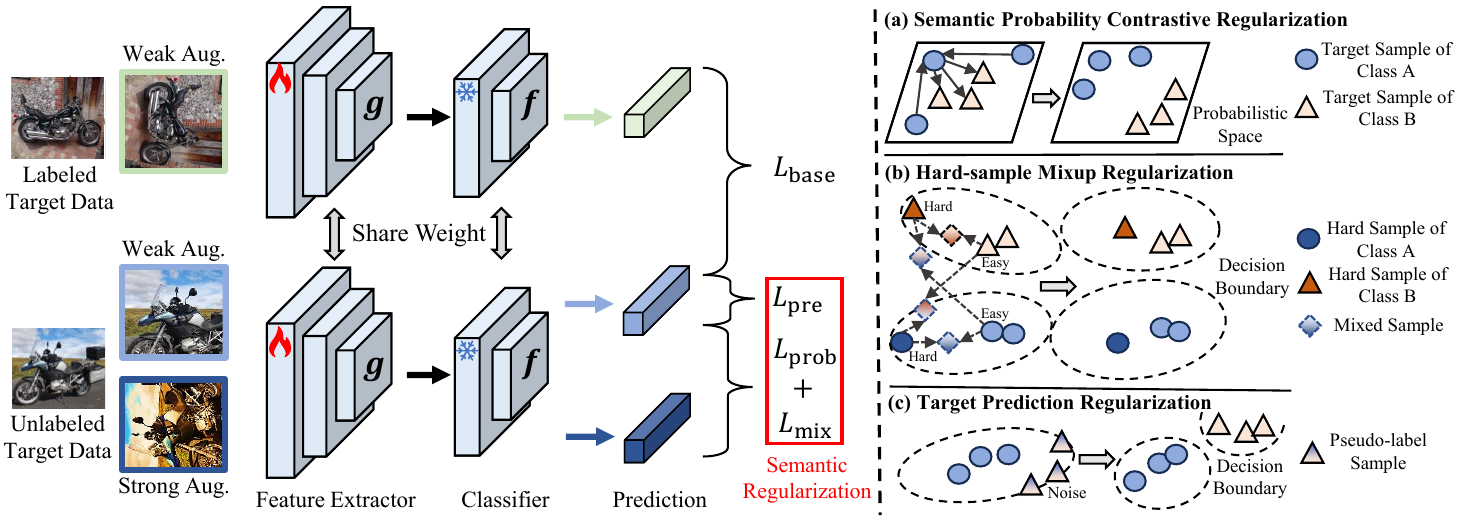}
	\caption{Illustration of our proposed semantic regularization learning (SERL) framework.
\textbf{Left:} The model initialized on the source domain is adaptively fine-tuned on the target domain.
The labeled target data and the strong and weak augmented versions of the unlabeled target data are input to the feature extractor $g$, then sent to the classifier $f$, and further learned the target domain knowledge through semantic regularization.
The two feature extractors and classifiers used share parameter weight.
\textbf{Right:} (a) Semantic probability contrastive regularization (SPCR) adaptively learns discriminative features through target semantic information and helps the model obtain a more confident probability output.
(b) Hard-sample mixup regularization (HMR) uses the semantic information of easy samples to guide the model in learning the distribution of hard target samples, helping the model learn more complex target domain distributions.
(c) Target prediction regularization (TPR) is used to minimize the misleading of erroneous semantic information to the model from the perspective of noise labels and help the model learn the true target domain distribution information.}
	\label{overview}
\end{figure*}

Following \cite{liang2020we,ma2022context}, we freeze $f(\cdot)$ and train $g(\cdot)$ when the model adapts to the target domain.
An overview of our SERL framework in the target adaptation stage is illustrated in Figure~\ref{overview}.
Following \cite{huang2023semi}, we generate the strong augment view for each unlabeled target sample $x_{i}^u$, represented as $\hat{x}_{i}^{u}$.
The target samples are then fed to the same feature extractor $g(\cdot)$ and classifier $f(\cdot)$ to obtain the probabilistic predictions $p_{i}^{u}$, $\hat{p} _{i}^{u}$, and the model is further adapted by the proposed semantic regularization learning.
For the labeled target data, we employ
the standard cross-entropy objective:
\begin{equation}
\begin{aligned}
L_\mathrm{l}=\frac1{N_{l}}\sum_{i=1}^{N_{l}}L_{CE}(p(y_i^{l}|x_i^{l}),y_i^{l}),
\end{aligned}
\end{equation}
where $L_{CE}$ is the standard cross-entropy loss.
For the unlabeled target data, we employ
the cross-entropy objective for its pseudo-label:
\begin{equation}
\begin{aligned}
L_\mathrm{u}=\frac1{N_{u}}\sum_{i=1}^{N_{u}}L_{CE}(p(y_i^{u}|x_i^{u}),y_i^{u}),
\end{aligned}
\end{equation}
where $y_i^{u}=\arg\max p_i^{u}$ is the pseudo-label of $x_i^u$.
Then, we utilize the mutual information maximization objective to encourage individually certain and globally diverse predictions:
\begin{equation}
L_\mathrm{mi}=\frac1{N_{u}}\sum_{i=1}^{N_{u}}\mathcal{H}\left(p(y_i^{u}|x_i^{u})\right)-\mathcal{H}\left(\sum_{i=1}^{N_{u}}p(y_i^{u}|x_i^{u})\right),
\end{equation}
where the entropy metric $\mathcal{H}(p(y|x))=\sum_{k=1}^cp_k\log p_k$ and $c$ is the number of different categories.

Following \cite{ma2022context}, we use a KNN-based pseudo-label propagation method.
In the neighbor graph, we can obtain one-hot pseudo-labels of unlabeled data through global propagation from labeled and low-uncertainty target data.
Finally, the base learning objective on the target domain can be derived as follows:
\begin{equation}
\label{base}
L_\mathrm{base}=L_\mathrm{l}+L_\mathrm{u}+L_\mathrm{mi}.
\end{equation}
On this basis, we will further introduce the proposed learning framework and how the training objective achieves further learning of the target domain through semantic probability contrastive regularization $L_\mathrm{prob}$, hard-sample mixup regularization $L_\mathrm{mix}$, and target prediction regularization $L_\mathrm{pre}$.

\subsection{Semantic Probability Contrastive Regularization}
After the model is initialized on the source domain, it will be fine-tuned on the target domain, and this process will not access the source domain data so that we can convert this semi-supervised domain adaptive process into a semi-supervised fine-tuning process for the target domain.
However, due to domain differences, the model still performs poorly on the target domain, even if it sees rich label information during the initialization of the source domain.

In recent years, contrastive learning \cite{oord2018representation,grill2020bootstrap,khosla2020supervised,chen2020simple,li2021probabilistic,huo2021heterogeneous,zhang2022semi, huang2024learningdifferentsamplessourcefree} has been proven to be an adequate representation learning method, which helps models better understand data and learn helpful knowledge representations in unsupervised or semi-supervised scenarios by constraining sample representation.
As a representative work, the self-supervised contrastive loss InfoNCE \cite{oord2018representation} takes the following format:
\begin{equation}
\label{infonce}
L_\mathrm{InfoNCE}=-\sum_{i=1}^{2N_u}\log\frac{\exp(z_i\cdot z_i^+/\tau)}{\sum_{j=1}^{2N_u}\mathbbm{1}_{(j\neq i)}\exp(z_i\cdot z_j/\tau)},
\end{equation}
where the $z_i^+$ represents the positive sample of feature embedding $z_i$, \added{$\mathbbm{1}_{(j\neq i)}$ represents the indicator function} and $\tau=0.15$ is the temperature coefficient.

In instance-based contrastive learning, two different augmented views from the same sample should be shown to represent similar features.
However, the knowledge learned only considering instance-level relationships is limited in complex target domains.
SupCon \cite{khosla2020supervised} learns more complex inter-sample relationships by introducing semantic information.
It is equivalent to applying semantic information regularization constraints to the model, which helps improve the generalization performance of the model on the target domain.
However, it is only applied to label-rich supervised learning, and feature-based contrastive learning cannot represent the actual target distribution of feature representations of many unlabeled target samples, which will impair the generalization ability of the classifier on the target domain \cite{li2021probabilistic}.
Inspired by \cite{li2021probabilistic,yang2022class}, we consider using semantic probability contrastive regularization based on adaptive weights to help the model better adapt to the target domain.
Specifically, we consider the following loss:
\begin{equation}
\label{prob}
L_\mathrm{prob}=-\sum_{i=1}^{2N_u}\sum_{k=1}^{2N_u}w_{ik}\log\frac{\exp(p_i^u\cdot p^{u+}_k/\tau)}{\sum_{j=1}^{2N_u}\mathbbm{1}_{j\neq i}\mathrm{exp}(p_i^u\cdot p_j^u/\tau)},
\end{equation}
where $p^{u+}_k$ represents the predicted probability of the positive target sample $k$ and the adaptive weight $w_{ik}$ is defined as follows:
\begin{equation}
\left.w_{ik}=\left\{
\begin{array}{ll}
1 & \text{if }k=i, \\
p_i^u\cdot p_k^u & \text{if }\arg\max p_i^u=\arg\max p_k^u, \\
0 & \text{otherwise}.
\end{array}
\right.\right.
\end{equation}

The adaptive weights give lower weights to low-confidence samples, which can help mitigate the impact of false constraints and reduce misunderstandings about the proper distribution of the target domain.

Compared with instance-level contrastive learning, we learn a more realistic target domain distribution by constraining the probability distribution of samples with the same semantic information.
At the same time, we constrain the similarity of the target samples from a probability perspective.
Specifically, for two samples $i$, $j$:
\begin{equation}
\begin{aligned}
p_i^u\cdot p_j^u=1 \Leftrightarrow &\arg\max(p_i^u)=\arg\max(p_j^u)\\
&\wedge\max(p_i^u)=\max(p_j^u)=1.
\label{10}
\end{aligned}
\end{equation}
\added{where $p_i^u\cdot p_j^u$ represents the product of two unsupervised samples, $\Leftrightarrow$ represents the equivalence, and $\wedge$ represents the logical AND relationship.
Eq. \ref{10} indicates that when optimizing $L_\mathrm{prob}$, the model forces the product between similarities to be maximized ($i.e.$, the product is 1), which is equivalent to the probability value corresponding to the predicted category ($i.e.$, $\arg\max(p_i^u)$) being 1.}
It encourages the prediction of the model to be close to the one-shot vector, $i.e.$, to make confident judgments on the target sample with the same semantic label, which helps to capture the semantic information on the target domain more effectively and has unique advantages in improving model performance.
Different from \cite{he2020momentum,chen2020simple}, we do not need a large batch size or sample queue to build comparison relationships, which can further save model memory consumption.

\subsection{Hard-sample Mixup Regularization}
Through semantic probability contrastive regularization, the model has been able to have a basic understanding of the sample relationships between target domains.
However, when we consider a more complex target domain relationship, $i.e.$, there are a certain number of complex samples on the target domain, which are usually distributed near the decision boundary and have low confidence, making it challenging to learn the complete target distribution further.
Existing SSDA methods ignore this problem, which makes them perform poorly in the face of complex target domain distributions.
Mixup is proven to reduce the overfitting tendency of the model by introducing a certain degree of regularization \cite{zhang2017mixup,berthelot2019mixmatch,zhang2020does,carratino2022mixup}.
Therefore, we can consider using Mixup to mix the semantic information of samples with different difficulty levels so that the model can better learn the complex semantic feature distribution of the target domain rather than adapt to the easy target distribution.

An important issue is partitioning samples with varying degrees of difficulty through existing models.
Previous work \cite{papyan2020prevalence} revealed that the weight vector of the trained last layer classifier converges to a high-dimensional geometric structure, which maximizes the separation of paired angles for all classifiers.
Another work \cite{ding2023proxymix} uses the weight vector of the classifier to construct pseudo-source domain samples to help model learning compensate for the lack of source domain knowledge.
Inspired by these works, we use the weight vectors of pre-trained classifiers on the source domain as anchors to divide easy and complex samples.
Specifically, we first define the classifier weight vectors $\{\textbf{c}_1,\textbf{c}_2,...,\textbf{c}_c\}$ of each category on the source domain as category anchors, search for and divide a certain number of easy and hard samples on unlabeled target domain data based on their distance from the anchors:
\begin{equation}
\begin{aligned}
x^{easy}_c&=\mathrm{argTOPK}\left(\min\left(dist\langle g(x^{u}),\textbf{c}_c\rangle\right), N^{easy}_u\right),
\end{aligned}
\end{equation}
\begin{equation}
\begin{aligned}
x^{hard}_c&=\mathrm{argTOPK}(\max\left(dist\langle g(x^{u}),\textbf{c}_c\rangle\right), N^{hard}_u),
\end{aligned}
\end{equation}
where $N^{easy}_u$ and $N^{hard}_u$ represent the number of easy and hard samples, \added{$\mathrm{argTOPK}(\cdot,N)$ means taking the first $N$ numbers, $\min(\cdot)$ and $\max(\cdot)$ mean sorting the objects from small to large/from large to small,} $dist\langle\cdot,\cdot\rangle$ represents the cosine distance between samples, and $x^{easy}_c$ and $x^{hard}_c$ represent the set of easy and hard samples for the $c$-th category.

To further enhance the understanding of semantic information, we connect easy and hard samples with their augmented versions $\hat{x}^{easy}$, $\hat{x}^{hard}$ to construct a vector represented as $X^{easy}=concate(x^{easy}, \hat{x}^{easy})$ and $X^{hard}=concate(x^{hard}, \hat{x}^{hard})$.
Furthermore, we will mix $X^{easy}$ and $X^{hard}$ to construct the following mixed training samples:
\begin{equation}
\begin{aligned}
X_i^{mix}&=\theta X_i^{easy}+(1-\theta)X_j^{hard},\\
y_i^{mix}&=\theta y_i^{easy}+(1-\theta)y_j^{hard},
\end{aligned}
\end{equation}
where $\theta$ is the mixup coefficient sampled from a random Beta distribution $Beta(\alpha,\alpha)$, $\alpha=1$. 
Following \cite{berthelot2019mixmatch}, we formulate the hard-sample mixup regularization loss as:
\begin{equation}
\label{mix}
L_\mathrm{mix}=\frac1{N_{u}}\sum_{i=1}^{N_{u}}\|f(g(X_i^{mix}))-y_i^{mix}\|_2^2,
\end{equation}
where $\|\cdot\|_2$ represents the $l_2$ regularization.

Unlike the cross-entropy loss, it is bound and more robust due to the sensitivity to corrupted labels.
The guidance of easy samples can help the model reduce the predicted distribution fluctuations between easy and hard samples when complex samples exhibit features more distinct from the source domain distribution \cite{xiong2023confidence}.
At the same time, this method imposes more complex semantic regularization constraints on the model, reducing the difference in semantic learning between easy and hard target samples, thereby helping the model better adapt to more complex target domain distributions.

\subsection{Target Prediction Regularization}
Our method relies on pseudo-labels generated by the model to form semantic information and use this as regularization information on the target domain without accessing source domain data during training.
Even if the learning of the model considers the discriminative knowledge of the target domain and hard sample information, the learning is still biased when there is much noise in the target pseudo-label.
Therefore, it is necessary to reduce the model from being misled by the semantic information generated by incorrect pseudo-labels.
However, existing SSDA methods ignore this impact, which will cause the model to generate noise due to domain shift and mislead the learning of clustering structures \cite{ huang2024learningdifferentsamplessourcefree}.
To reduce the misguidance brought by erroneous semantic information to the model, we minimize the impact of erroneous pseudo-labels from the perspective of prediction regularization and further help the model learn correct target domain distribution knowledge.

Inspired by \cite{yi2023source,bai2021understanding,song2019prestopping,liu2020early}, we exploit the early training phenomenon to address the potential spurious label noise problem.
Specifically, the early training phenomenon shows that classifiers can predict mislabeled samples with relatively high accuracy in the early adaptation stage before memorizing mislabeled target data.
To leverage predictions made during early training, we employ early learning regularization (ELR), encouraging model predictions to adhere to early sample predictions.
The regularization term is given by:
\begin{equation}
\label{elr}
L_\mathrm{pre}=\frac1{N_{u}}\sum_{i=1}^{N_{u}}\log(1-\tilde{y}^{ut\top}_{i} p_i^{ut}),
\end{equation}
where $p_i^{ut}$ is the target probability output at epoch $t$, $\tilde{y}_i^{ut} = \beta \tilde{y}_i^{u(t-1)}+(1-\beta)p_i^{ut}$ is the moving average prediction and $\beta=0.7$ is the hyper-parameter.

Note that minimizing Eq. \ref{elr} forces $p_i^{ut}$ to be close to $\tilde{y}^{ut}_i$.
Therefore, Eq. \ref{elr} prevents the model from remembering target label noise by forcing the model predictions to stay close to the moving average predictions $\tilde{y}^{ut}$ of these most likely accurate target labels, further reducing the impact of noisy semantic information on the model brought about misguidance.
Combined with all the components mentioned above, the whole algorithm of our SERL can be described using Algorithm \ref{alg}.

\begin{algorithm}[t]
    \caption{SERL Framework for SSDA.}
    \label{alg}
    \renewcommand{\algorithmicrequire}{\textbf{Input:}}
    \renewcommand{\algorithmicensure}{\textbf{Initialization:}}
    \begin{algorithmic}[1]
        \REQUIRE Labeled source data $\{x^s_i, y^s_i\}^{N_s}_{i=1}$, labeled target data $\{x^l_i, y^l_i\}^{N_l}_{i=1}$, unlabeled target set $\{x^u_i\}^{N_u}_{i=1}$ and strongly augmented unlabeled target set $\{\hat{x}^u_i\}^{N_u}_{i=1}$. The number of training epochs \({T}\). The trade-off hyper-parameters $\lambda_\mathrm{prob}$, $\lambda_\mathrm{mix}$, and $\lambda_\mathrm{pre}$. 
        \ENSURE Freeze the final classifier layer $f$, and copy the parameters from the source feature extractor to the target feature extractor as initialization.
        \STATE \textcolor{blue}{\textit{\# Adaptive on Target Domain}} \\
        \FOR{each $t \in [1, T]$}
            \STATE Compute self-supervised pseudo-labels for $x^u_i$.
        \FOR{each target data  $x^u_i\in [1, N_u]$}
            \STATE \textcolor{blue}{\textit{\# Calculate Losses}} \\
            \STATE Compute the base loss $L_\mathrm{base}$ with Eq.~\ref{base};\\
            \STATE Compute the probability contrastive loss $L_\mathrm{prob}$ with Eq.~\ref{prob};\\
            \STATE Compute the mixup loss $L_\mathrm{mix}$ with Eq.~\ref{mix};\\
            \STATE Compute the prediction regularization loss $L_\mathrm{pre}$ with Eq.~\ref{elr};\\
            \STATE \textcolor{blue}{\textit{\# Parameter Optimization.}} \\
            \STATE Update the parameters in target feature extractor $g$ via $L_\mathrm{all}$ in Eq.~\ref{allloss}.
        \ENDFOR
        \ENDFOR
        \RETURN The updated parameters of the target feature extractor $g$.
    \end{algorithmic}
\end{algorithm}

\begin{table*}[t]
\centering
\caption{Accuracy (\%) on \textit{\textit{DomainNet}} under the settings of 1-shot and ResNet-34 as backbone networks.}
\renewcommand\arraystretch{1.1}
\tabcolsep=9.8pt{
\begin{tabular}{c|c|ccccccc|c}
\toprule
{SF} & {Method} & {R$\rightarrow$C} & {R$\rightarrow$P} & {P$\rightarrow$C} & {C$\rightarrow$S} & {S$\rightarrow$P} & {R$\rightarrow$S} & {P$\rightarrow$R} & {Mean}\\
\midrule
$\times$ &S+T&55.6&60.6&56.8&50.8&56.0&46.3&71.8&56.9\\
$\times$&DANN~\cite{ganin2016domain}&58.2&61.4&56.3&52.8&57.4&52.2&70.3&58.4\\
$\times$&ENT~\cite{grandvalet2004semi}&65.2&65.9&65.4&54.6&59.7&52.1&75.0&62.6\\
$\times$&MME~\cite{saito2019semi}&70.0&67.7&69.0&56.3&64.8&61.0&76.1&66.4\\
$\times$&UODA~\cite{qin2021contradictory}&72.7&70.3&69.8&60.5&66.4&62.7&77.3&68.5\\
$\times$&BiAT~\cite{jiang2020bidirectional}&73.0&68.0&71.6&57.9&63.9&58.5&77.0&67.1\\
$\times$&APE~\cite{kim2020attract}&70.4&70.8&72.9&56.7&64.5&63.0&76.6&67.6\\
$\times$&STar~\cite{singh2021improving} & 74.1 & 71.3 & 71.0 & 63.5 & 66.1 & 64.1 & 80.0 & 70.0 \\ 
$\times$&DECOTA~\cite{yang2021deep}&79.1&74.9&76.9&65.1&72.0&69.7&79.6&73.9\\
$\times$&CDAC~\cite{li2021cross}&77.4&74.2&75.5&67.6&71.0&69.2&80.4&73.6\\
$\times$&CLDA~\cite{singh2021clda}&76.1&75.1&71.0&63.7&70.2&67.1&80.1&71.9\\
$\times$&ECACL~\cite{li2021ecacl}&75.3&74.1&75.3&65.0&72.1&68.1&79.7&72.8\\
$\times$&ASDA~\cite{qin2022semi}&77.0&75.4&75.5&66.5&72.1&70.9&79.7&73.9\\
$\times$&MCL~\cite{yan2022multi}&77.4&74.6&75.5&66.4&74.0&70.7&82.0&74.4\\
$\times$&ProML~\cite{huang2023semi}&78.5&75.4&77.8&70.2&74.1&72.4&84.0&76.1\\
$\times$&SLA~\cite{yu2023semi}&79.8&75.6&77.4&68.1&71.7&71.7&80.4&75.0\\
$\times$&IDMNE~\cite{li2024inter}&79.6&76.0&79.4&71.7&75.4&73.5&82.1&76.8\\
$\times$&G-ABC~\cite{li2023adaptive}& 80.7&76.8&79.3&72.0&75.0& 73.2& 83.4 & 77.5 \\ 
\hline
\checkmark&DEEM~\cite{ma2022context} & 79.7 & 78.1 & 77.0 & 71.9 & 77.7 & 76.7 & 85.4 & 78.1 \\ 
\checkmark&SERL (Ours)&\textbf{90.5}&\textbf{88.8}&\textbf{90.2}&\textbf{89.1}&\textbf{90.1}&\textbf{87.1}&\textbf{93.3}&\textbf{89.9}\\  
\bottomrule
\end{tabular}}
\label{domainnet1}
\end{table*}

\begin{table*}[t]
\centering
\caption{Accuracy (\%) on \textit{\textit{DomainNet}} under the settings of 3-shot and ResNet-34 as backbone networks.}
\renewcommand\arraystretch{1.1}
\tabcolsep=9.8pt{
\begin{tabular}{c|c|ccccccc|c}
\toprule
{SF}&{Method} & {R$\rightarrow$C} & {R$\rightarrow$P} & {P$\rightarrow$C} & {C$\rightarrow$S} & {S$\rightarrow$P} & {R$\rightarrow$S} & {P$\rightarrow$R} & {Mean}\\
\midrule
$\times$ & S+T&60.0&62.2&59.4&55.0&59.5&50.1&73.9&60.0\\
$\times$&DANN~\cite{ganin2016domain}&59.8&62.8&59.6&55.4&59.9&54.9&72.2&60.7\\
$\times$&ENT~\cite{grandvalet2004semi}&71.0&69.2&71.1&60.0&62.1&61.1&78.6&67.6\\
$\times$&MME~\cite{saito2019semi}&72.2&69.7&71.7&61.8&66.8&61.9&78.5&68.9\\
$\times$&UODA~\cite{qin2021contradictory}&75.4&71.5&73.2&64.1&69.4&64.2&80.8&71.2\\
$\times$&BiAT~\cite{jiang2020bidirectional}&74.9&68.8&74.6&61.5&67.5&62.1&78.6&69.7\\
$\times$&APE~\cite{kim2020attract}&76.6&72.1&76.7&63.1&66.1&67.8&79.4&71.7\\
$\times$&STar~\cite{singh2021improving} & 77.1 & 73.2 & 75.8 & 67.8 & 69.2 & 67.9& 81.2& 73.2 \\ 
$\times$&DECOTA~\cite{yang2021deep}&80.4&75.2&78.7&68.6&72.7&71.9&81.5&75.6\\ 
$\times$&CDAC~\cite{li2021cross}&79.6&75.1&79.3&69.9&73.4&72.5&81.9&76.0\\
$\times$&CLDA~\cite{singh2021clda}&77.7&75.7&76.4&69.7&73.7&71.1&82.9&75.3\\
$\times$&ECACL~\cite{li2021ecacl}&79.0&77.3&79.4&70.6&74.6&71.6&82.4&76.4\\
$\times$&ASDA~\cite{qin2022semi}&79.4&76.7&78.3&70.2&74.2&72.1&82.3&76.2\\
$\times$&MCL~\cite{yan2022multi}&79.4&76.3&78.8&70.9&74.7&72.3&83.3&76.5\\
$\times$&ProML~\cite{huang2023semi}&80.2&76.5&78.9&72.0&75.4&73.5&84.8&77.4\\
$\times$&SLA~\cite{yu2023semi}&81.6&76.0&80.3&71.3&73.5&73.5&82.5&76.9\\
$\times$&IDMNE~\cite{li2024inter}&80.8&76.9&80.3&73.2&75.4&73.9&82.8&77.5\\
$\times$&G-ABC~\cite{li2023adaptive}& 82.1& 76.7& 81.6& 73.7& 76.3& 74.3& 83.9& 78.2 \\
\hline
\checkmark&DEEM~\cite{ma2022context} & 80.5 & 79.0 & 77.5& 74.9 & 80.0 & 75.9 & 88.5 & 79.5 \\
\checkmark&SERL (Ours)&\textbf{91.8}&\textbf{89.1}&\textbf{91.9}&\textbf{89.9}&\textbf{92.1}&\textbf{87.5}&\textbf{94.3}&\textbf{90.9}\\
\bottomrule
\end{tabular}}
\label{domainnet3}
\end{table*}

\begin{table*}[ht]
\centering
\caption{Accuracy (\%) on \textit{Office-Home} under the settings of 1-shot using VGGNet-16 as the backbone network.}
\renewcommand\arraystretch{1.2}
\tabcolsep=2.8pt
\begin{tabular}{c|c|cccccccccccc|c}
\toprule
SF & Method   & R$\rightarrow$C   & R$\rightarrow$P  & R$\rightarrow$A    & P$\rightarrow$R    & P$\rightarrow$C  & P$\rightarrow$A      & A$\rightarrow$P  & A$\rightarrow$C  & A$\rightarrow$R  & C$\rightarrow$R   & C$\rightarrow$A   & C$\rightarrow$P  & Mean  \\ 
\midrule
$\times$&S+T & 39.5 & 75.3 & 61.2 & 71.6 & 37.0 & 52.0 & 63.6 & 37.5 & 69.5 & 64.5 & 51.4 & 65.9 & 57.4 \\ 
$\times$&DANN~\cite{ganin2016domain} & 52.0 & 75.7 & 62.7 & 72.7 & 45.9 & 51.3 & 64.3 & 44.4 & 68.9 & 64.2 & 52.3 & 65.3 & 60.0 \\ 
$\times$&ENT~\cite{grandvalet2004semi} & 23.7 & 77.5 & 64.0 & 74.6 & 21.3 & 44.6 & 66.0 & 22.4 & 70.6 & 62.1 & 25.1 & 67.7 & 51.6 \\ 
$\times$&MME~\cite{saito2019semi} & 49.1 & 78.7 & 65.1 & 74.4 & 46.2 & 56.0 & 68.6 & 45.8 & 72.2 & 68.0 & 57.5 & 71.3 & 62.7 \\ 
$\times$&UODA~\cite{qin2021contradictory} & 49.6 & 79.8 & 66.1 & 75.4 & 45.5 & 58.8 & 72.5 & 43.3 & 73.3 & 70.5 & 59.3 & 72.1 & 63.9 \\ 
$\times$&DECOTA~\cite{yang2021deep} & 47.2 & 80.3 & 64.6 &  75.5 & 47.2& 56.6 & 71.1& 42.5 & 73.1 & 71.0 & 57.8 & 72.9 & 63.3 \\
$\times$&ASDA~\cite{qin2022semi} & 51.6 & 80.9 & 66.9 & 75.9 & 49.7 & 60.5 & 71.0 & 44.9 & 73.2 & 70.6 & 58.7 & 72.8 & 64.7 \\
$\times$&IDMNE~\cite{li2024inter}& 52.6& 81.8& 67.5& 77.3& 50.7& 59.7& 73.7& 49.6& 72.6& 71.4& 62.5 & 76.2 & 66.3 \\
\hline
\checkmark&DEEM~\cite{ma2022context} & 62.5 & 82.1 & 68.5 & 79.0 & 62.1& 65.4& 76.5& 60.3 & 76.1 & 74.6 & 63.3 & 75.4 & 70.5 \\ 
\checkmark&SERL (Ours)& \textbf{74.4}& \textbf{92.8}& \textbf{78.0}& \textbf{89.4}& \textbf{70.6} & \textbf{72.2} & \textbf{86.7} & \textbf{74.7} & \textbf{86.1}& \textbf{84.3} & \textbf{72.7}  & \textbf{86.8}&\textbf{80.6}\\
\bottomrule
\end{tabular}
\label{office-home1}
\end{table*}

\begin{table*}[ht]
\centering
\caption{Accuracy (\%) on \textit{Office-Home} under the settings of 3-shot using VGGNet-16 as the backbone network.}
\renewcommand\arraystretch{1.2}
\tabcolsep=2.8pt
\begin{tabular}{c|c|cccccccccccc|c}
\toprule
SF & Method    & R$\rightarrow$C   & R$\rightarrow$P    & R$\rightarrow$A   & P$\rightarrow$R   & P$\rightarrow$C     & P$\rightarrow$A    & A$\rightarrow$P   & A$\rightarrow$C   & A$\rightarrow$R  & C$\rightarrow$R  & C$\rightarrow$A    & C$\rightarrow$P   & Mean  \\ 
\midrule
$\times$&S+T & 49.6 & 78.6 & 63.6 & 72.7 & 47.2 & 55.9 & 69.4 & 47.5 & 73.4 & 69.7 & 56.2 & 70.4 & 62.9 \\ 
$\times$&DANN~\cite{ganin2016domain} & 56.1 & 77.9 & 63.7 & 73.6 & 52.4 & 56.3 & 69.5 & 50.0 & 72.3 & 68.7 & 56.4 & 69.8 & 63.9 \\ 
$\times$&ENT~\cite{grandvalet2004semi} & 48.3 & 81.6 & 65.5 & 76.6 & 46.8 & 56.9 & 73.0 & 44.8 & 75.3 & 72.9 & 59.1 & 77.0 & 64.8 \\ 
$\times$&MME~\cite{saito2019semi} & 56.9 & 82.9 & 65.7 & 76.7 & 53.6 & 59.2 & 75.7 & 54.9 & 75.3 & 72.9 & 61.1 & 76.3 & 67.6 \\ 
$\times$&UODA~\cite{qin2021contradictory} & 57.6 & 83.6 & 67.5 & 77.7 & 54.9 & 61.0 & 77.7 & 55.4 & 76.7 & 73.8 & 61.9 & 78.4 & 68.9 \\ 
$\times$&APE~\cite{kim2020attract}& 56.0 & 81.0 & 65.2 & 73.7 & 51.4 & 59.3 & 75.0 & 54.4 & 73.7 & 71.4 & 61.7 & 75.1 & 66.5 \\ 
$\times$&DECOTA~\cite{yang2021deep} & 59.9 & 83.9 & 67.7 & 77.3 & 57.7& 60.7 & 78.0& 54.9 & 76.0 &  74.3& 63.2 & 78.4 & 69.3 \\
$\times$&ASDA~\cite{qin2022semi} & 59.3 & 83.6 & 68.0 & 78.3 & 56.8 & 61.8 & 78.6 & 55.7 & 75.3 & 74.0 & 63.3 & 78.9 & 69.5 \\
$\times$&IDMNE~\cite{li2024inter}& 60.2& 84.4& 69.3& 77.9& 59.2& 62.6& 77.7& 58.2& 76.7& 74.9& 64.6& 79.3 & 70.4 \\
\hline
\checkmark&DEEM~\cite{ma2022context} & 69.3 & 86.6 & 69.8 & 79.3 & 66.3& 64.0& 80.1 & 64.0 & 77.8 & 75.6 & 63.7 & 78.3 & 72.9 \\
\checkmark&SERL (Ours)& \textbf{79.6}   & \textbf{92.8}   & \textbf{78.4}  & \textbf{90.0}    & \textbf{78.3}   & \textbf{72.8}  & \textbf{90.1}   & \textbf{78.4}    & \textbf{86.8}   & \textbf{89.6}  & \textbf{74.2}    & \textbf{91.5}   & \textbf{83.5}\\
\bottomrule
\end{tabular}
\label{office-home3}
\end{table*}

\begin{table*}[t]
\centering
\caption{Accuracy (\%) on \textit{Office-31} under the settings of 1-shot and 3-shot using AlexNet as the backbone network.}
\renewcommand\arraystretch{1.2}
\tabcolsep=10.8pt
\begin{tabular}{c|c|ccc|ccc}
    \hline
     \multirow{2}{*}{SF}&\multirow{2}{*}{Method} & \multicolumn{3}{c|}{1-shot}  & \multicolumn{3}{c}{3-shot} \\ 
    \cline{3-8}\xrowht{5pt}
    
    ~ & ~ & D$\rightarrow$A & W$\rightarrow$A & Mean & D$\rightarrow$A & W$\rightarrow$A & Mean   \\ 
    \hline
    $\times$&  S+T & 50.0 & 50.4 & 50.2 & 62.4 & 61.2  & 61.8   \\ 
    $\times$&   DANN~\cite{ganin2016domain} & 54.5 & 57.0 & 55.8 & 65.2 & 64.4  & 64.8  \\ 
    $\times$&    ENT~\cite{grandvalet2004semi} &50.0  &50.7  &50.4  &66.2  &64.0  &65.1 \\
   $\times$&     MME~\cite{saito2019semi} & 55.8 & 57.2 & 56.5 & 67.8 & 67.3  & 67.6   \\ 
    $\times$&    BiAT~\cite{jiang2020bidirectional} & 54.6 & 57.9 & 56.3 & 68.5 & 68.2  & 68.3  \\ 
    $\times$&    APE~\cite{kim2020attract} & - & - & - & 67.6 & 69.0  & 68.3  \\
    $\times$&    CLDA~\cite{singh2021clda} & 62.7 & 64.6 & 63.6 & 72.5 & 70.5 & 71.5  \\ 
    $\times$&    CDAC~\cite{li2021cross} & 62.8 & 63.4 & 63.1 & 70.0 & 70.1  & 70.0  \\ 
    $\times$&    STar~\cite{singh2021improving} & 56.8 & 59.8 & 58.3 & 69.0 & 69.1  & 69.1  \\ 
    $\times$&    IDMNE~\cite{li2024inter} & - & - & - & 71.3 & 71.0  & 71.2  \\
   $\times$&     G-ABC~\cite{li2023adaptive} & 65.7 & 67.9 & 66.8 & 73.1 & 71.0 & 72.0 \\ 
   \hline
    \checkmark&  DEEM~\cite{ma2022context} & 75.7 & 76.6 & 76.2 & 76.8 & 78.5 & 77.7   \\
    \checkmark&    SERL (Ours)& \textbf{79.0}& \textbf{81.1} & \textbf{80.1} & \textbf{82.1} & \textbf{82.5} & \textbf{82.3}\\
\hline
\end{tabular}
\label{office-31}
\end{table*}

\section{EXPERIMENT}
\label{experiment}
\subsection{Datasets}
We evaluate our proposed method on three widely used datasets, including DomainNet \cite{peng2019moment}, Office-Home \cite{venkateswara2017deep}, and Office-31 \cite{saenko2010adapting}.
For fairness of comparison, we have one or three samples on the target domain during training for each category in different datasets.

\textbf{DomainNet} is a significant benchmark dataset designed to evaluate multi-source domain adaptation methods composed of 345 classes, six domains: Clipart, Infographics, Painting, Real, Sketch, and Quickdraw, and each domain contains 126 image categories.
Similar to MME~\cite{saito2019semi}, we use a subset of the DomainNet as one of our evaluation benchmarks.
We only select four domains: Real (R), Clipart (C), Painting (P), and Sketch (S), because other domains and categories may contain excessive sample noise.
Following MME~\cite{saito2019semi}, we conduct adaptation experiments on seven scenarios on these four domains.

\textbf{Office-Home} is a medium-sized SSDA benchmark dataset with many challenging object recognition domain adaptation scenarios.
It consists of four domains: Art (A), Clipart (C), Products (P), and Real (R).
The dataset contains images of 65 object classes typically constructed in office and home environments for each domain.
We consider 12 domain adaptation scenarios compared with previous SSDA methods to achieve a fair comparison.

\textbf{Office-31} is a small dataset containing three domains: Amazon (A), DSLR (D), and Webcam (W), with 31 categories on each domain.
Following MME~\cite{saito2019semi}, we choose Amazon (A) as the target domain because compared to Webcam (W) and DSLR (D), each category in Amazon has sufficiently rich samples.
Therefore, we only consider two adaptation scenarios on this small SSDA dataset: "W$\rightarrow$A” and “D$\rightarrow$A.”

\subsection{Implementation Details}
We select three feature extraction backbones, including AlexNet \cite{krizhevsky2012imagenet}, VGGNet-16 \cite{simonyan2014very}, and ResNet-34 \cite{he2016deep} with pre-trained weights on ImageNet \cite{krizhevsky2012imagenet}. 
Similar to \cite{liang2020we,ma2022context}, for AlexNet and VGGNet-16, we add a bottleneck layer after the last layer of the feature extractor. We then use a classifier with a normalized, fully connected layer.
For ResNet-34, we remove the last layer of the feature extractor, add a bottleneck layer like the previous backbone network, and use a classifier with fully connected layers.
We randomly select three mini-batches from $N_s$, $N_l$, and $N_u$ during each iteration.
For batch sizes, they are 64, 32, and 64 for AlexNet, 32, 16, and 32 for VGGNet-16, and 48, 24, and 48 for ResNet-34.
The learning rates of the feature extractor, bottleneck layer, and classifier are set to 0.001, 0.01, and 0.01, respectively, and the weight decay is 0.0005. 
The loss weights $\lambda_\mathrm{prob}$, $\lambda_\mathrm{mix}$, and $\lambda_\mathrm{pre}$ are specified as 0.3, 60, and 3, respectively. 
The number of easy samples $N^{easy}_u$ and hard samples $N^{hard}_u$ are 15.
We adopt the widely used Randaugmnt \cite{cubuk2020randaugment} as the strong data augmentation strategy.
Our experiments were implemented using Pytorch \cite{paszke2019pytorch} and run on an RTX 3090 GPU.
We use three different randomized seeds to obtain fairer experimental results.

\subsection{Comparison With State-of-the-Arts}
In this section, we compare the classification performance of our proposed SERL method with previous state-of-the-art SSDA algorithms, including S+T, DANN~\cite{ganin2016domain}, ENT~\cite{grandvalet2004semi}, MME~\cite{saito2019semi}, UODA~\cite{qin2021contradictory}, BiAT~\cite{jiang2020bidirectional}, APE~\cite{kim2020attract}, STar~\cite{singh2021improving}, DECOTA~\cite{yang2021deep}, ECACL~\cite{li2021ecacl}, ASDA~\cite{qin2022semi}, MCL~\cite{yan2022multi}, SLA~\cite{yu2023semi}, CLDA~\cite{singh2021clda}, CDAC~\cite{li2021cross}, ProML~\cite{huang2023semi}, DEEM~\cite{ma2022context}, IDMNE~\cite{li2024inter}, G-ABC~\cite{li2023adaptive}.
Note that S+T refers to the method of training an adaptive model by supervising only labeled samples from two domains,
DANN~\cite{ganin2016domain} applies standard cross-entropy loss to SSDA by using it to some labeled samples on the target domain.
SLA~\cite{yu2023semi} is a plug-and-play SSDA method, and we consider combining it with CDAC~\cite{li2021cross}, the best result reported in their paper.

\subsubsection{Results on DomainNet}
Tables \ref{domainnet1} and \ref{domainnet3} present the quantitative comparison results of our proposed method with numerous existing alternatives on the DomainNet benchmark.
For the large dataset DomainNet, we use 1-shot and 3-shot settings and ResNet-34 with a relatively deep network structure as the corresponding backbone network.
It can be seen from the results that our method outperforms all previous methods in all scenarios on 1-shot and 3-shot settings and achieves enormous advantages.
Specifically, SERL improves the previous best-performing SSDA algorithm DEEM in the 1-shot and 3-shot settings of all adaptive scenarios, respectively, with the average accuracy increased by 11.8\% and 11.4\%.
It is worth noting that DEEM is also based on the source-free SSDA method, but our performance is better, which is all attributed to our semantic regularization learning method.
Most of the existing methods use the source-with training paradigm. 
Compared with them, we have improved the average accuracy of G-ABC by 12.4\% and 12.7\% in the 1-shot and 3-shot settings, respectively.
By comparing the two tables, we can find that the performance in the 1-shot setting is slightly inferior to the improvement in the 3-shot setting.
This means that our method requires more supervision to realize its potential better since more labeled target examples help better to learn the semantic information of the target domain.

\subsubsection{Results on Office-Home}
To further validate the feasibility of the proposed SERL framework in SSDA scenarios, Tables \ref{office-home1} and \ref{office-home3} present the quantitative results and comparison of our method in benchmark Office-Home compared to previous methods.
We conducted experiments on the dataset using VGGNet-16 as the backbone network in 1-shot and 3-shot settings and all 12 Office-Home adaptation scenarios.
It is worth noting that our method outperforms all existing methods in all scenarios and significantly outperforms the source-free SSDA method DEEM by 10.1\% for 1-shot and 10.6\% for 3-shot and the source-with method IDMNE by 14.3\% for 1-shot and 13.1\% for 3-shot in terms of average accuracy, further demonstrating the superiority of our method.

\subsubsection{Results on Office-31}
Table \ref{office-31} shows the results of our comparison with existing methods on Office-31.
Office-31 is a small dataset, and in order to maintain consistency with existing methods, we use AlexNet with a relatively small number of layers to conduct experiments under 1-shot and 3-shot.
It can be seen from the results that the average accuracy of our method under the 1-shot setting is 80.1\%, and the average accuracy under the 3-shot setting is 82.3\%, respectively surpassing the existing state-of-the-art SSDA method DEEM 3.9\% and 4.6\%.
Compared with DomainNet and Office-Home, its performance improvement is relatively limited.
This is because Office-31 contains a few images and is a relatively simple SSDA dataset.
In contrast, DomainNet and Office-Home have richer image data, providing more challenging environments and room for improvement.
This shows that our method is more capable of handling more complex domain adaptation scenarios than existing methods, proving the superiority of the proposed method on SSDA tasks.

\begin{center}
\begin{table}[t]
\centering
\caption{Accuracy (\%) of ablation study on \textit{DomainNet} under the settings of 1-shot with the ResNet-34 backbone.}
\renewcommand\arraystretch{1.1}
\tabcolsep=8.0pt
\scalebox{0.9}{
\begin{tabular}{c|cccc|cc|c}
    \hline
    Num. & $L_\mathrm{base}$ & $L_\mathrm{prob}$ & $L_\mathrm{mix}$ & $L_\mathrm{pre}$ & R$\rightarrow$C & R$\rightarrow$P & Mean  \\ 
    \hline 
        1 & \added{\checkmark}& &  &  & \added{79.0} & \added{77.8} & \added{78.4}\\
        2 & \checkmark&\checkmark &  &  & 86.4 & 83.9 & 85.2 \\ 
        3 & \checkmark& & \checkmark &  & 81.1 & 79.8 & 80.5 \\ 
        4 &  \checkmark& &  & \checkmark & 83.7 & 81.9 & 82.8 \\ 
        5 & \checkmark&\checkmark & \checkmark &  & 87.4 & 85.8 & 86.6 \\ 
        6 & \checkmark&\checkmark &  & \checkmark & 88.9 & 87.6 & 88.3 \\ 
        7 &  \checkmark& & \checkmark & \checkmark & 84.3 & 83.1 & 83.7 \\ 
        8 & \checkmark &\checkmark & \checkmark & \checkmark & \textbf{90.5} & \textbf{88.8} & \textbf{89.7}\\
    \hline
    \end{tabular}}
\label{absMainComponent}
\end{table}
\end{center}

\subsection{Ablation Study}
\subsubsection{Each Main Component}
We conducted ablation studies on the main components in 1-shot and 3-shot settings for DomainNet R$\rightarrow$C and R$\rightarrow$P, as shown in Table \ref{absMainComponent}.
Rows \added{2-4} show that each component can produce significant improvements.
Rows \added{5-7} show that each combination still improves performance, indicating the versatility of the proposed module.
At the same time, the SPCR module and the TPR module can bring more significant improvements to the model than the HMR module.
This is because, in the SPCR and TPR modules, the model has learned good feature representations for most samples on the target domain, resulting in a limited number of potentially hard samples, so the improvement is relatively limited.
The best performance is achieved when all components of the model are activated.
\vspace{-1.8em}

\begin{center}
\begin{table}[t]
\centering
\caption{Accuracy (\%) of ablation study for Probability Contrast and Adaptive Weight in SPCR with 1-shot setting.}
\renewcommand\arraystretch{1.1}
\tabcolsep=4.0pt
\begin{tabular}{cc|ccc}
    \hline\xrowht{20pt}
        \makecell[c]{Probability \\ Contrast} & \makecell[c]{Adaptive \\ Weight} & \makecell[c]{Office-Home \\ R$\rightarrow$P} & \makecell[c]{DomainNet \\ C$\rightarrow$S} & \makecell[c]{Mean} \\ \hline
         &  & 84.0 & 83.1&83.6 \\ 
        \checkmark & & 88.6 & 87.6&88.1 \\ 
        \checkmark & \checkmark & \textbf{92.8} & \textbf{89.1}& \textbf{91.0}\\ \hline
    \end{tabular}
\label{ProbabilityContrastandAdaptiveWeight}
\end{table}
\end{center}

\subsubsection{Source-Free Learning Framework}
\begin{figure}[t]
	\centering
 \includegraphics[width=0.9\linewidth]{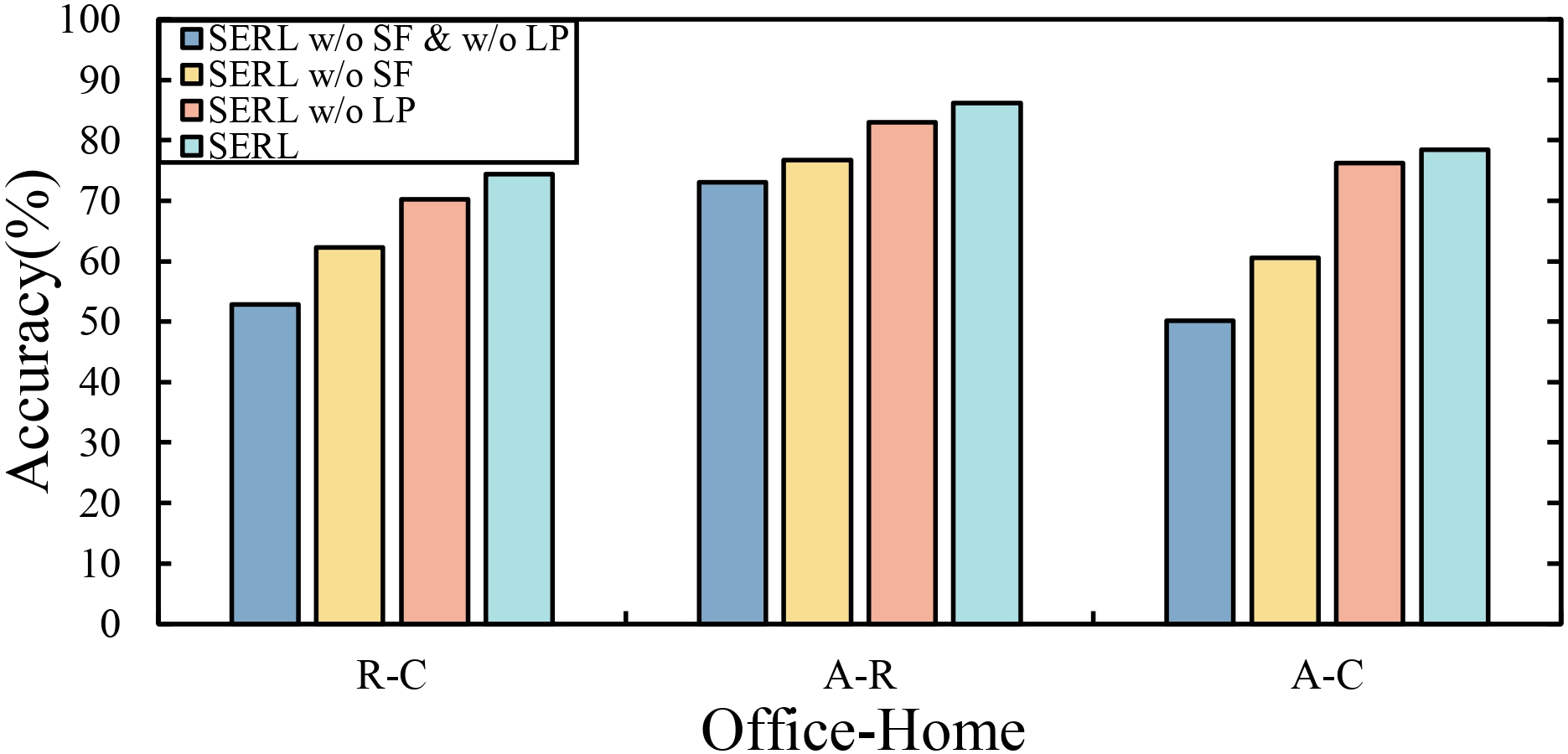}
	\caption{The impact of source-free learning frameworks on performance.
The experiments were conducted in three scenarios of \textit{Office-Home} under the 1-shot setting.
\textit{SF} stands for source-free training paradigm, and \textit{LP} stands for label propagation.}
	\label{SourceFreeLearningFramework}
\end{figure}

\begin{figure*}[t]
    \centering
	  \includegraphics[width=1\linewidth]{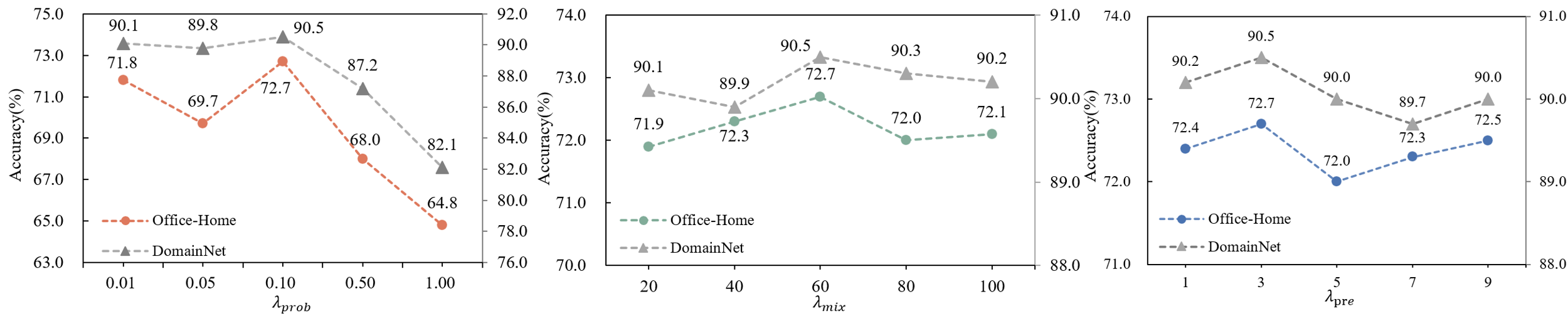}
	  \caption{The effect of different loss balance parameters $\lambda_\mathrm{prob}$, $\lambda_\mathrm{mix}$, and $\lambda_\mathrm{pre}$ on the model classification accuracy in the \textit{Office-Home} C$\rightarrow$A and \textit{DomainNet} R$\rightarrow$C scenario under the 1-shot setting.}
	  \label{senmainloss} 
      \vspace{-1.5em}
\end{figure*}

\begin{figure}[t]
	\centering
 \includegraphics[width=0.8\linewidth]{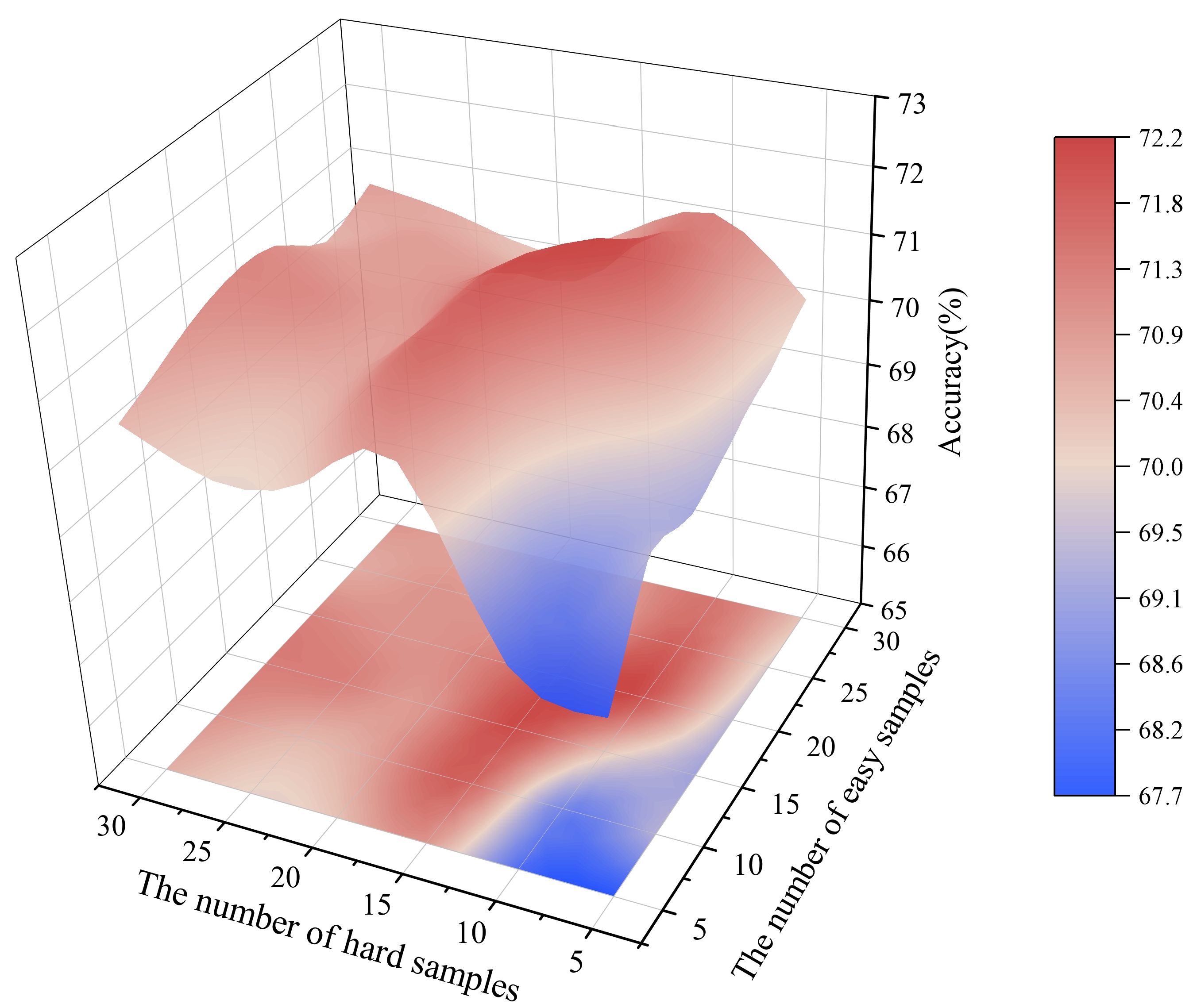}
	\caption{Variation in model performance for different numbers of easy and hard samples $N^{easy}_u, N^{hard}_u \in \{5, 10, 15, 20, 25, 30\}$ for the 1-shot setting in the P$\rightarrow$A scenario of the \textit{Office-Home} dataset.}
	\label{easyhardsample}
\end{figure}

To prove the importance of the source-free training framework, we show the ablation experimental results in different cross-domain scenarios of Office-Home in Figure \ref{SourceFreeLearningFramework}.
When source-free training strategies and label propagation methods are not considered, the performance of the model will drop to the lowest point.
This shows that the source-free training framework can allow the model to focus on learning a more accurate target domain distribution, thereby significantly improving the performance of the model.
\added{Since the target domain has only a small amount of labeled data, while the source domain has a large amount of labeled data for supervision.
The number of this part of supervision signals creates a strong contrast between the source domain and the target domain.
When there are only a few labeled data, the model can easily rely on the characteristics of the source domain to make decisions.}
When considering either alone, the performance of the model drops significantly compared to the performance of the complete model.
In particular, the source-free training method can bring more significant performance improvement to the model.
This is because source-free only considers fine-tuning the source model on the target domain, which can reduce the impact of source domain samples on the adaptation of the target domain during training, allowing the model to focus more on learning semantic information on the target domain.

\subsubsection{Probability Contrast and Adaptive Weight in SPCR}
We investigated specific techniques mentioned in SPCR to prove the effectiveness of our SPCR further, as shown in Table \ref{ProbabilityContrastandAdaptiveWeight}.
It is worth noting that when nothing is considered, the model degrades to the InfoNCE loss, as shown in Eq. \ref{infonce}.
When considering learning discriminative features from the probability space, the model performance improves significantly because the model is forced to output more confident representation information and can be combined with the knowledge learned by the classifier to allow the feature extractor to learn more compact target representation clusters.
When considering adding adaptive weights, the model can adaptively learn relevant target representations for objects of the same category with different confidence levels, thereby achieving the best performance.

\subsection{Further Analysis}
\subsubsection{Sensitivity of \texorpdfstring{$\lambda_\mathrm{prob}$}{}, \texorpdfstring{$\lambda_\mathrm{mix}$}{} and \texorpdfstring{$\lambda_\mathrm{pre}$}{}}
We show the impact of the loss balance parameters $\lambda_\mathrm{prob}$, $\lambda_\mathrm{mix}$ and $\lambda_\mathrm{pre}$ on the classification accuracy under the Office-Home C$\rightarrow$A scenario in Figure \ref{senmainloss}.
It can be observed that when $\lambda_\mathrm{prob}=0.1$, $\lambda_\mathrm{mix}=60$, and $\lambda_\mathrm{pre}=3$, the trained model achieves the highest performance in image classification.

\begin{figure}[t]
	\centering
 \includegraphics[width=1\linewidth]{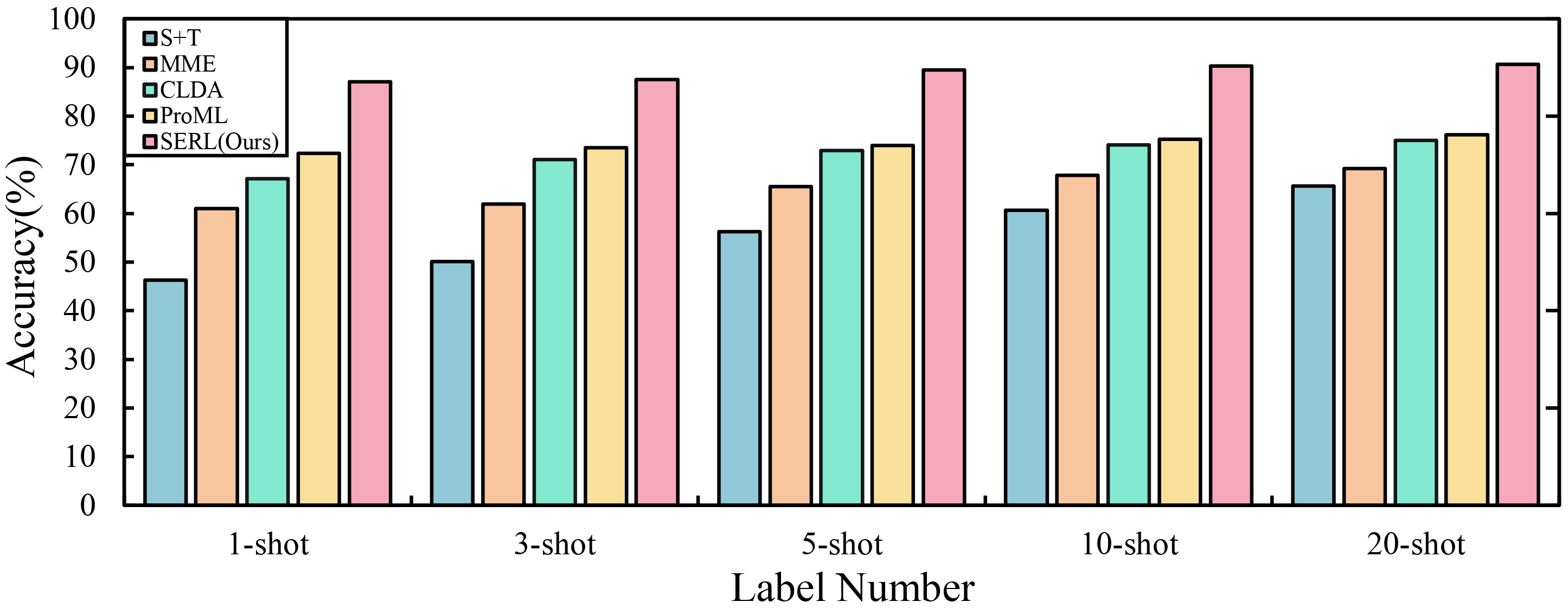}
	\caption{Histogram of quantitative comparison under different number of labeled samples settings on \textit{DomainNet} R$\rightarrow$S.}
	\label{SensitivityofLabeleSamples}
\end{figure}

\begin{figure}[t]
    \centering
	  \subfloat[Office-31 W$\rightarrow$A]{\includegraphics[width=0.48\linewidth]{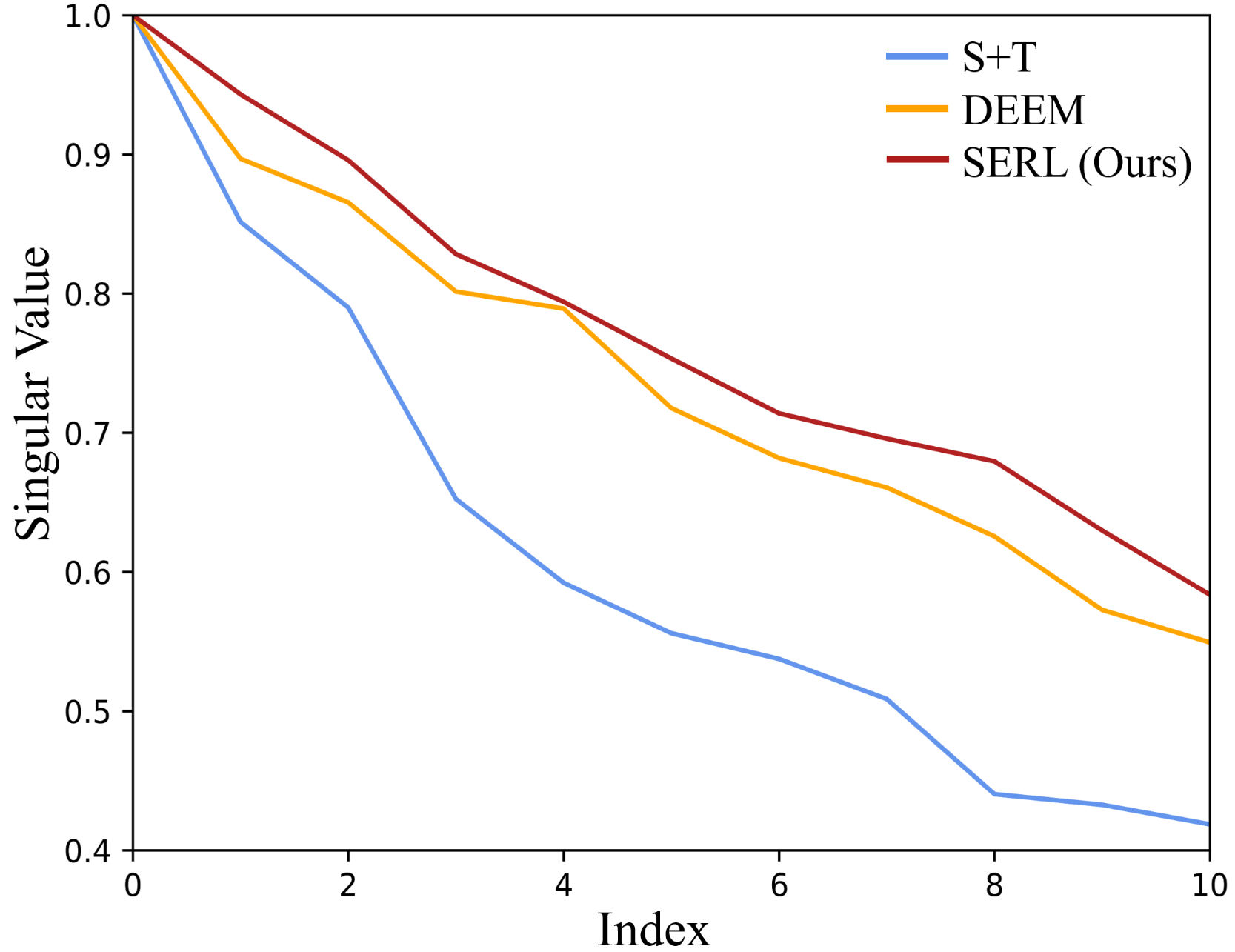}}
        \hspace{0mm}
	  \subfloat[Office-31 D$\rightarrow$A]{\includegraphics[width=0.482\linewidth]{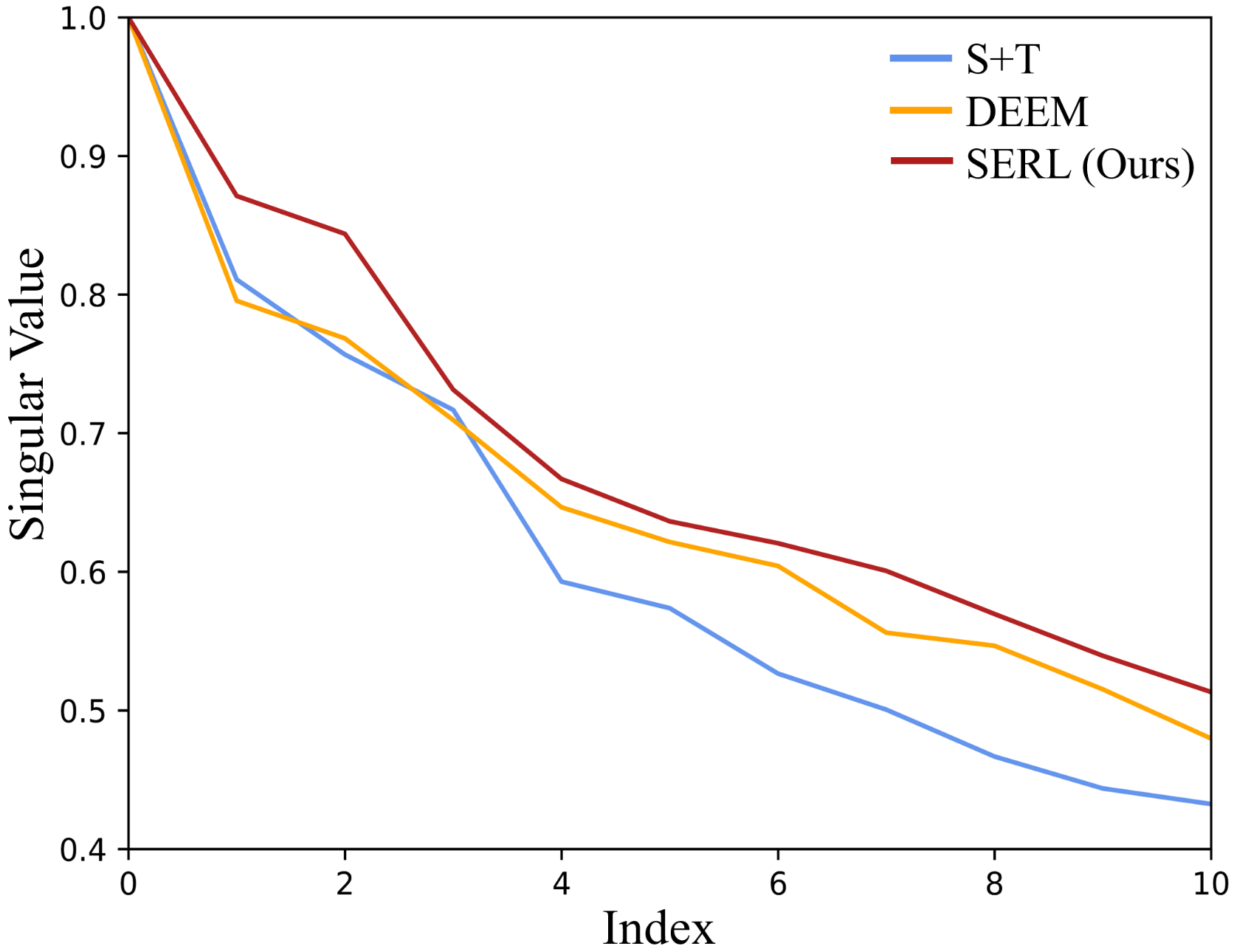}}
	  \caption{The SVD analysis of feature matrices obtained by different methods in different 1-shot scenarios.}
	  \label{svd} 
      \vspace{-1.5em}
\end{figure}

\subsubsection{Sensitivity of \texorpdfstring{$N^{easy}_u$}{} and \texorpdfstring{$N^{hard}_u$}{} in HMR}
Regarding the number of easy and hard samples we mentioned in HMR, $i.e.$, $N^{easy}_u$ and $N^{hard}_u$, we further analyze its impact on model performance, as shown in Figure \ref{easyhardsample}.
It can be seen from the results that the blue part is mainly concentrated in areas with a small number of samples, which shows that the model performs poorly when the number of mixed samples is small.
The model performance improves when the number of mixed samples is gradually increased.
This proves that the model can learn new target domain semantic representations by gradually adding the number of target samples.

\begin{figure*}[t]
	\centering
 \includegraphics[width=1\linewidth]{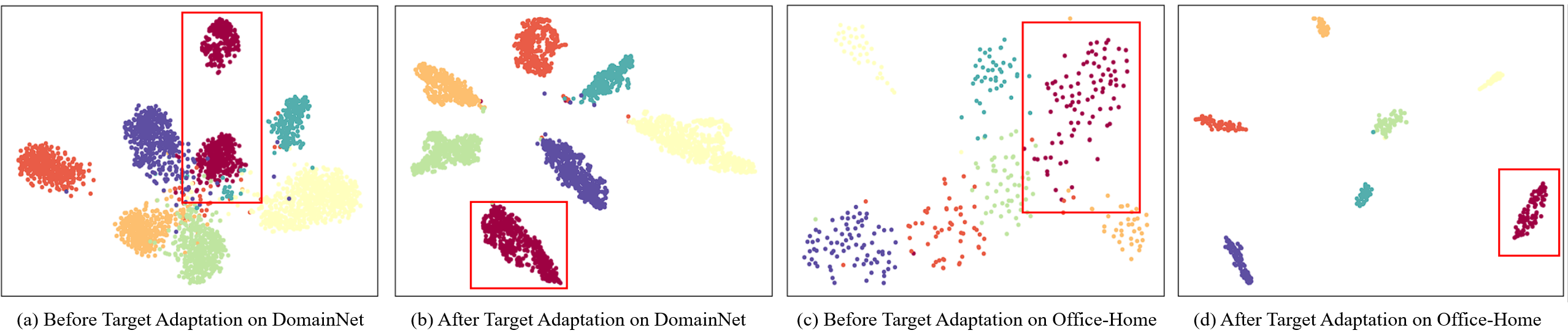}
	\caption{Feature visualization using t-SNE \cite{van2008visualizing}.
We randomly selected seven categories and assigned them different colors for the 3-shot scenes of \textit{DomainNet} R$\rightarrow$S and \textit{Office-Home} C$\rightarrow$P.
\added{The red box shows obvious differences.}}
	\label{tsne}
    \vspace{-1.5em}
\end{figure*}

\begin{figure}[t]
    \centering
	  \subfloat[Easy Samples]{\includegraphics[width=0.45\linewidth]{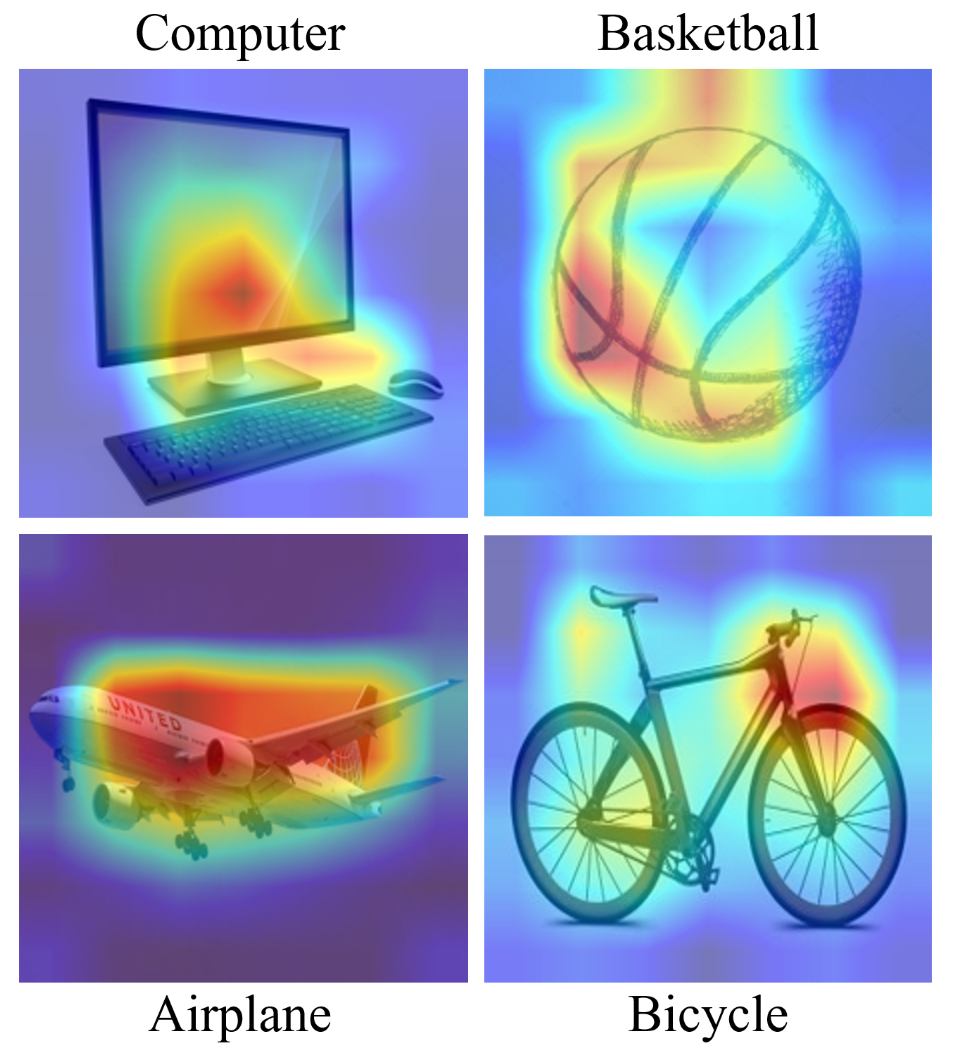}}
        \hspace{5mm}
	  \subfloat[Hard Samples]{\includegraphics[width=0.456\linewidth]{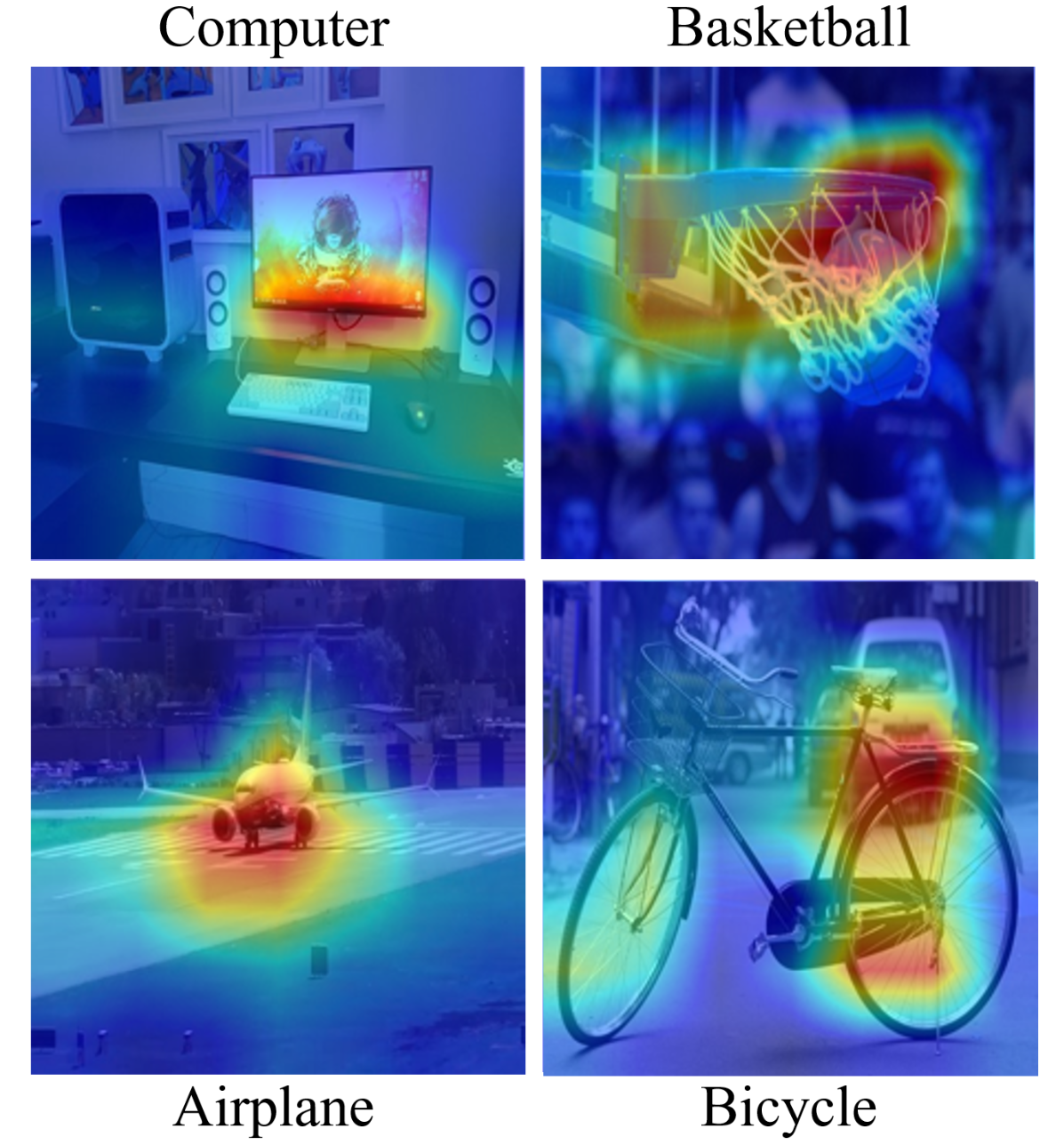}}
	  \caption{The Grad-CAM \cite{selvaraju2017grad} visualization of the features generated by our SERL for different samples in the \textit{DomainNet} dataset.}
	  \label{gradcam} 
      \vspace{-1.5em}
\end{figure}

\subsubsection{Sensitivity of Labeled Samples}
In Figure \ref{SensitivityofLabeleSamples}, we show histograms comparing our method with existing methods under different labeled samples.
Our method still maintains optimal performance even with more labeled data.
At the same time, as the number of labels increases, the improvement of methods gradually decreases.
This phenomenon suggests diminishing returns to more labels, eventually leading to a fully supervised learning situation.

\subsubsection{Spectral Analysis}
To further analyze the discriminability of the learned features, following \cite{chen2019transferability,xie2022collaborative}, we perform singular value decomposition (SVD) analysis on the feature matrices extracted under the 1-shot setting for the Office-31 W$\rightarrow$A and D$\rightarrow$A scenario.
The results are shown in Figure \ref{svd}.
Relative to SERL, the largest singular values of the feature matrices of S+T and DEEM are significantly larger than the other singular values, greatly weakening the information signal of the feature vectors corresponding to smaller singular values.
Such a sharp distribution of singular values implies a deterioration of distinguishability.
However, the singular values of the feature matrices learned by our proposed SERL successfully reduce the large difference between the largest value and the remaining values while maintaining higher values, which implies that more dimensions corresponding to smaller singular values positively affect the classification and intuitively improve the discriminability of the features.

\subsection{Feature Visualization}
\subsubsection{Feature Aggregation}
As shown in Figure \ref{tsne}, we use t-SNE \cite{van2008visualizing} to visualize the changes in deep features during training.
For DomainNet R$\rightarrow$S, where the domain difference is relatively small, the model trained only on the source domain can better aggregate most of the same features. 
However, it performs poorly in Office-Home C$\rightarrow$P, where the domain difference is relatively significant.
However, as training proceeds, learned features from different domains belonging to the same class are mapped nearby and clustered together, while those from different classes are clearly separated, and the clusters are more evenly distributed.
The results show that using the proposed SERL can produce domain-invariant and differentiated target features, helping the model perform well on the target domain.

\subsubsection{Attention Visualization}
In Figure \ref{gradcam}, we use the Grad-CAM \cite{selvaraju2017grad} to visualize the attention maps of the model for different categories of target samples in the DomainNet dataset after target domain adaptation.
Whether the model faces easy samples with relatively easy backgrounds or hard samples with relatively complex backgrounds, the model can capture the key information of the target samples, which is due to the assistance of our SERL for the model to learn the semantic information on the target domain.

\section{Conclusion}
\label{conclusion}
This paper proposes a novel SSDA learning framework called semantic regularization learning (SERL), which provides regularization constraints by learning semantic information from the target data, thereby better learning the representation distribution of the target domain.
This paper considers fine-tuning the feature extractor on the target domain based on the source pre-trained model.
Firstly, semantic probability contrastive regularization helps the model learn more discriminative feature representations, using semantic information on the target domain to understand the similarities and differences between samples.
At the same time, it encourages the model to make confident judgments, helping to capture the semantic information on the target domain more fully.
Then, hard-sample mixup regularization is proposed to learn more complex target domains by reducing the fluctuation of predictive distributions between easy and hard samples through a guidance strategy for easy samples.
Finally, target prediction regularization corrects erroneous target predictions by maximizing the correlation between the prediction output and the early learned target, reducing the misleading of false semantic information.
Extensive experiments and comprehensive analysis with good performance on three benchmark datasets demonstrate the superiority of our method, which significantly surpasses existing methods and achieves impressive results.

\ifCLASSOPTIONcaptionsoff
  \novelpage
\fi

\bibliographystyle{ieeetr}
\bibliography{bibtex/bib/IEEEabrv,bibtex/bib/IEEEexample}

\begin{thebibliography}{10}

\bibitem{krizhevsky2012imagenet}
A.~Krizhevsky, I.~Sutskever, and G.~E. Hinton, ``Imagenet classification with deep convolutional neural networks,'' {\em Advances in neural information processing systems}, vol.~25, 2012.

\bibitem{he2016deep}
K.~He, X.~Zhang, S.~Ren, and J.~Sun, ``Deep residual learning for image recognition,'' in {\em Proceedings of the IEEE conference on computer vision and pattern recognition}, pp.~770--778, 2016.

\bibitem{rastegari2016xnor}
M.~Rastegari, V.~Ordonez, J.~Redmon, and A.~Farhadi, ``Xnor-net: Imagenet classification using binary convolutional neural networks,'' in {\em European conference on computer vision}, pp.~525--542, Springer, 2016.

\bibitem{krizhevsky2017imagenet}
A.~Krizhevsky, I.~Sutskever, and G.~E. Hinton, ``Imagenet classification with deep convolutional neural networks,'' {\em Communications of the ACM}, vol.~60, no.~6, pp.~84--90, 2017.

\bibitem{qiao2019transductive}
L.~Qiao, Y.~Shi, J.~Li, Y.~Wang, T.~Huang, and Y.~Tian, ``Transductive episodic-wise adaptive metric for few-shot learning,'' in {\em Proceedings of the IEEE/CVF international conference on computer vision}, pp.~3603--3612, 2019.

\bibitem{yan2016image}
Y.~Yan, F.~Nie, W.~Li, C.~Gao, Y.~Yang, and D.~Xu, ``Image classification by cross-media active learning with privileged information,'' {\em IEEE Transactions on Multimedia}, vol.~18, no.~12, pp.~2494--2502, 2016.

\bibitem{wang2016csps}
J.~Wang, W.~Wang, R.~Wang, and W.~Gao, ``Csps: An adaptive pooling method for image classification,'' {\em IEEE Transactions on Multimedia}, vol.~18, no.~6, pp.~1000--1010, 2016.

\bibitem{long2015fully}
J.~Long, E.~Shelhamer, and T.~Darrell, ``Fully convolutional networks for semantic segmentation,'' in {\em Proceedings of the IEEE conference on computer vision and pattern recognition}, pp.~3431--3440, 2015.

\bibitem{gao2022fbsnet}
G.~Gao, G.~Xu, J.~Li, Y.~Yu, H.~Lu, and J.~Yang, ``Fbsnet: A fast bilateral symmetrical network for real-time semantic segmentation,'' {\em IEEE Transactions on Multimedia}, 2022.

\bibitem{yan2019semantic}
B.~Yan, X.~Niu, B.~Bare, and W.~Tan, ``Semantic segmentation guided pixel fusion for image retargeting,'' {\em IEEE Transactions on Multimedia}, vol.~22, no.~3, pp.~676--687, 2019.

\bibitem{minaee2021image}
S.~Minaee, Y.~Boykov, F.~Porikli, A.~Plaza, N.~Kehtarnavaz, and D.~Terzopoulos, ``Image segmentation using deep learning: A survey,'' {\em IEEE transactions on pattern analysis and machine intelligence}, vol.~44, no.~7, pp.~3523--3542, 2021.

\bibitem{li2023muva}
Z.~Li, W.~Ye, J.~Terven, Z.~Bennett, Y.~Zheng, T.~Jiang, and T.~Huang, ``Muva: A new large-scale benchmark for multi-view amodal instance segmentation in the shopping scenario,'' in {\em Proceedings of the IEEE/CVF International Conference on Computer Vision}, pp.~23504--23513, 2023.

\bibitem{wang2023seggpt}
X.~Wang, X.~Zhang, Y.~Cao, W.~Wang, C.~Shen, and T.~Huang, ``Seggpt: Segmenting everything in context,'' {\em arXiv preprint arXiv:2304.03284}, 2023.

\bibitem{pan2010domain}
S.~J. Pan, I.~W. Tsang, J.~T. Kwok, and Q.~Yang, ``Domain adaptation via transfer component analysis,'' {\em IEEE transactions on neural networks}, vol.~22, no.~2, pp.~199--210, 2010.

\bibitem{patel2015visual}
V.~M. Patel, R.~Gopalan, R.~Li, and R.~Chellappa, ``Visual domain adaptation: A survey of recent advances,'' {\em IEEE signal processing magazine}, vol.~32, no.~3, pp.~53--69, 2015.

\bibitem{mei2020instance}
K.~Mei, C.~Zhu, J.~Zou, and S.~Zhang, ``Instance adaptive self-training for unsupervised domain adaptation,'' in {\em Computer Vision--ECCV 2020: 16th European Conference, Glasgow, UK, August 23--28, 2020, Proceedings, Part XXVI 16}, pp.~415--430, Springer, 2020.

\bibitem{zhang2021self}
C.~Zhang, Z.~Li, J.~Liu, P.~Peng, Q.~Ye, S.~Lu, T.~Huang, and Y.~Tian, ``Self-guided adaptation: Progressive representation alignment for domain adaptive object detection,'' {\em IEEE Transactions on Multimedia}, vol.~24, pp.~2246--2258, 2021.

\bibitem{ganin2015unsupervised}
Y.~Ganin and V.~Lempitsky, ``Unsupervised domain adaptation by backpropagation,'' in {\em International conference on machine learning}, pp.~1180--1189, PMLR, 2015.

\bibitem{deng2021informative}
W.~Deng, L.~Zhao, Q.~Liao, D.~Guo, G.~Kuang, D.~Hu, M.~Pietik{\"a}inen, and L.~Liu, ``Informative feature disentanglement for unsupervised domain adaptation,'' {\em IEEE Transactions on Multimedia}, vol.~24, pp.~2407--2421, 2021.

\bibitem{wang2022cross}
R.~Wang, Z.~Wu, Z.~Weng, J.~Chen, G.-J. Qi, and Y.-G. Jiang, ``Cross-domain contrastive learning for unsupervised domain adaptation,'' {\em IEEE Transactions on Multimedia}, 2022.

\bibitem{jing2022adversarial}
M.~Jing, L.~Meng, J.~Li, L.~Zhu, and H.~T. Shen, ``Adversarial mixup ratio confusion for unsupervised domain adaptation,'' {\em IEEE Transactions on Multimedia}, 2022.

\bibitem{lu2021discriminative}
Y.~Lu, D.~Li, W.~Wang, Z.~Lai, J.~Zhou, and X.~Li, ``Discriminative invariant alignment for unsupervised domain adaptation,'' {\em IEEE Transactions on Multimedia}, vol.~24, pp.~1871--1882, 2021.

\bibitem{zhao2020review}
S.~Zhao, X.~Yue, S.~Zhang, B.~Li, H.~Zhao, B.~Wu, R.~Krishna, J.~E. Gonzalez, A.~L. Sangiovanni-Vincentelli, S.~A. Seshia, {\em et~al.}, ``A review of single-source deep unsupervised visual domain adaptation,'' {\em IEEE Transactions on Neural Networks and Learning Systems}, vol.~33, no.~2, pp.~473--493, 2020.

\bibitem{zuo2023dual}
Y.~Zuo, H.~Yao, L.~Zhuang, and C.~Xu, ``Dual structural knowledge interaction for domain adaptation,'' {\em IEEE Transactions on Multimedia}, no.~99, pp.~1--15, 2023.

\bibitem{ding2023unsupervised}
F.~Ding, J.~Li, W.~Tian, S.~Zhang, and W.~Yuan, ``Unsupervised domain adaptation via risk-consistent estimators,'' {\em IEEE Transactions on Multimedia}, 2023.

\bibitem{saito2019semi}
K.~Saito, D.~Kim, S.~Sclaroff, T.~Darrell, and K.~Saenko, ``Semi-supervised domain adaptation via minimax entropy,'' in {\em Proceedings of the IEEE/CVF International Conference on Computer Vision}, pp.~8050--8058, 2019.

\bibitem{li2021cross}
J.~Li, G.~Li, Y.~Shi, and Y.~Yu, ``Cross-domain adaptive clustering for semi-supervised domain adaptation,'' in {\em Proceedings of the IEEE/CVF Conference on Computer Vision and Pattern Recognition}, pp.~2505--2514, 2021.

\bibitem{li2021ecacl}
K.~Li, C.~Liu, H.~Zhao, Y.~Zhang, and Y.~Fu, ``Ecacl: A holistic framework for semi-supervised domain adaptation,'' in {\em Proceedings of the IEEE/CVF International Conference on Computer Vision}, pp.~8578--8587, 2021.

\bibitem{qin2022semi}
C.~Qin, L.~Wang, Q.~Ma, Y.~Yin, H.~Wang, and Y.~Fu, ``Semi-supervised domain adaptive structure learning,'' {\em IEEE Transactions on Image Processing}, vol.~31, pp.~7179--7190, 2022.

\bibitem{xu2022semi}
H.-M. Xu, L.~Liu, Q.~Bian, and Z.~Yang, ``Semi-supervised semantic segmentation with prototype-based consistency regularization,'' {\em Advances in neural information processing systems}, 2022.

\bibitem{yan2022multi}
Z.~Yan, Y.~Wu, G.~Li, Y.~Qin, X.~Han, and S.~Cui, ``Multi-level consistency learning for semi-supervised domain adaptation,'' {\em arXiv preprint arXiv:2205.04066}, 2022.

\bibitem{yu2023semi}
Y.-C. Yu and H.-T. Lin, ``Semi-supervised domain adaptation with source label adaptation,'' in {\em Proceedings of the IEEE/CVF Conference on Computer Vision and Pattern Recognition}, pp.~24100--24109, 2023.

\bibitem{huang2023semi}
X.~Huang, C.~Zhu, and W.~Chen, ``Semi-supervised domain adaptation via prototype-based multi-level learning,'' {\em arXiv preprint arXiv:2305.02693}, 2023.

\bibitem{li2023adaptive}
J.~Li, G.~Li, and Y.~Yu, ``Adaptive betweenness clustering for semi-supervised domain adaptation,'' {\em IEEE Transactions on Image Processing}, 2023.

\bibitem{li2024inter}
J.~Li, G.~Li, and Y.~Yu, ``Inter-domain mixup for semi-supervised domain adaptation,'' {\em Pattern Recognition}, vol.~146, p.~110023, 2024.

\bibitem{chen2023semi}
T.~Chen, Y.~Guo, S.~Hao, and R.~Hong, ``Semi-supervised domain adaptation for major depressive disorder detection,'' {\em IEEE Transactions on Multimedia}, 2023.

\bibitem{kim2020attract}
T.~Kim and C.~Kim, ``Attract, perturb, and explore: Learning a feature alignment network for semi-supervised domain adaptation,'' in {\em European conference on computer vision}, pp.~591--607, Springer, 2020.

\bibitem{liang2020we}
J.~Liang, D.~Hu, and J.~Feng, ``Do we really need to access the source data? source hypothesis transfer for unsupervised domain adaptation,'' in {\em International conference on machine learning}, pp.~6028--6039, PMLR, 2020.

\bibitem{xuan2020hard}
H.~Xuan, A.~Stylianou, X.~Liu, and R.~Pless, ``Hard negative examples are hard, but useful,'' in {\em Computer Vision--ECCV 2020: 16th European Conference, Glasgow, UK, August 23--28, 2020, Proceedings, Part XIV 16}, pp.~126--142, Springer, 2020.

\bibitem{zuo2021challenging}
L.~Zuo, M.~Jing, J.~Li, L.~Zhu, K.~Lu, and Y.~Yang, ``Challenging tough samples in unsupervised domain adaptation,'' {\em Pattern Recognition}, vol.~110, p.~107540, 2021.

\bibitem{liu2022complementary}
Y.~Liu, H.~Ge, L.~Sun, and Y.~Hou, ``Complementary attention-driven contrastive learning with hard-sample exploring for unsupervised domain adaptive person re-id,'' {\em IEEE Transactions on Circuits and Systems for Video Technology}, vol.~33, no.~1, pp.~326--341, 2022.

\bibitem{xiong2023confidence}
Y.~Xiong, H.~Chen, Z.~Lin, S.~Zhao, and G.~Ding, ``Confidence-based visual dispersal for few-shot unsupervised domain adaptation,'' in {\em Proceedings of the IEEE/CVF International Conference on Computer Vision}, pp.~11621--11631, 2023.

\bibitem{zhang2017mixup}
H.~Zhang, M.~Cisse, Y.~N. Dauphin, and D.~Lopez-Paz, ``mixup: Beyond empirical risk minimization,'' {\em arXiv preprint arXiv:1710.09412}, 2017.

\bibitem{liu2020early}
S.~Liu, J.~Niles-Weed, N.~Razavian, and C.~Fernandez-Granda, ``Early-learning regularization prevents memorization of noisy labels,'' {\em Advances in neural information processing systems}, vol.~33, pp.~20331--20342, 2020.

\bibitem{yi2023source}
L.~Yi, G.~Xu, P.~Xu, J.~Li, R.~Pu, C.~Ling, A.~I. McLeod, and B.~Wang, ``When source-free domain adaptation meets learning with noisy labels,'' {\em arXiv preprint arXiv:2301.13381}, 2023.

\bibitem{pei2018multi}
Z.~Pei, Z.~Cao, M.~Long, and J.~Wang, ``Multi-adversarial domain adaptation,'' in {\em Thirty-second AAAI conference on artificial intelligence}, 2018.

\bibitem{venkateswara2017deep}
H.~Venkateswara, J.~Eusebio, S.~Chakraborty, and S.~Panchanathan, ``Deep hashing network for unsupervised domain adaptation,'' in {\em Proceedings of the IEEE Conference on Computer Vision and Pattern Recognition}, pp.~5018--5027, 2017.

\bibitem{saenko2010adapting}
K.~Saenko, B.~Kulis, M.~Fritz, and T.~Darrell, ``Adapting visual category models to new domains,'' in {\em Computer Vision--ECCV 2010: 11th European Conference on Computer Vision, Heraklion, Crete, Greece, September 5-11, 2010, Proceedings, Part IV 11}, pp.~213--226, Springer, 2010.

\bibitem{gretton2012kernel}
A.~Gretton, K.~M. Borgwardt, M.~J. Rasch, B.~Sch{\"o}lkopf, and A.~Smola, ``A kernel two-sample test,'' {\em The Journal of Machine Learning Research}, vol.~13, no.~1, pp.~723--773, 2012.

\bibitem{ganin2016domain}
Y.~Ganin, E.~Ustinova, H.~Ajakan, P.~Germain, H.~Larochelle, F.~Laviolette, M.~Marchand, and V.~Lempitsky, ``Domain-adversarial training of neural networks,'' {\em The journal of machine learning research}, vol.~17, no.~1, pp.~2096--2030, 2016.

\bibitem{long2017deep}
M.~Long, H.~Zhu, J.~Wang, and M.~I. Jordan, ``Deep transfer learning with joint adaptation networks,'' in {\em International conference on machine learning}, pp.~2208--2217, PMLR, 2017.

\bibitem{sun2017correlation}
B.~Sun, J.~Feng, and K.~Saenko, ``Correlation alignment for unsupervised domain adaptation,'' {\em Domain adaptation in computer vision applications}, pp.~153--171, 2017.

\bibitem{zhuo2017deep}
J.~Zhuo, S.~Wang, W.~Zhang, and Q.~Huang, ``Deep unsupervised convolutional domain adaptation,'' in {\em Proceedings of the 25th ACM international conference on Multimedia}, pp.~261--269, 2017.

\bibitem{tzeng2017adversarial}
E.~Tzeng, J.~Hoffman, K.~Saenko, and T.~Darrell, ``Adversarial discriminative domain adaptation,'' in {\em Proceedings of the IEEE conference on computer vision and pattern recognition}, pp.~7167--7176, 2017.

\bibitem{zhang2021joint}
B.~Zhang, T.~Chen, B.~Wang, and R.~Li, ``Joint distribution alignment via adversarial learning for domain adaptive object detection,'' {\em IEEE Transactions on Multimedia}, vol.~24, pp.~4102--4112, 2021.

\bibitem{xie2018learning}
S.~Xie, Z.~Zheng, L.~Chen, and C.~Chen, ``Learning semantic representations for unsupervised domain adaptation,'' in {\em International conference on machine learning}, pp.~5423--5432, PMLR, 2018.

\bibitem{shen2018wasserstein}
J.~Shen, Y.~Qu, W.~Zhang, and Y.~Yu, ``Wasserstein distance guided representation learning for domain adaptation,'' in {\em Proceedings of the AAAI Conference on Artificial Intelligence}, vol.~32, 2018.

\bibitem{ge2023unsupervised}
P.~Ge, C.-X. Ren, X.-L. Xu, and H.~Yan, ``Unsupervised domain adaptation via deep conditional adaptation network,'' {\em Pattern Recognition}, vol.~134, p.~109088, 2023.

\bibitem{shermin2020adversarial}
T.~Shermin, G.~Lu, S.~W. Teng, M.~Murshed, and F.~Sohel, ``Adversarial network with multiple classifiers for open set domain adaptation,'' {\em IEEE Transactions on Multimedia}, vol.~23, pp.~2732--2744, 2020.

\bibitem{chen2019progressive}
C.~Chen, W.~Xie, W.~Huang, Y.~Rong, X.~Ding, Y.~Huang, T.~Xu, and J.~Huang, ``Progressive feature alignment for unsupervised domain adaptation,'' in {\em Proceedings of the IEEE/CVF conference on computer vision and pattern recognition}, pp.~627--636, 2019.

\bibitem{pan2019transferrable}
Y.~Pan, T.~Yao, Y.~Li, Y.~Wang, C.-W. Ngo, and T.~Mei, ``Transferrable prototypical networks for unsupervised domain adaptation,'' in {\em Proceedings of the IEEE/CVF conference on computer vision and pattern recognition}, pp.~2239--2247, 2019.

\bibitem{zhong2021does}
L.~Zhong, Z.~Fang, F.~Liu, J.~Lu, B.~Yuan, and G.~Zhang, ``How does the combined risk affect the performance of unsupervised domain adaptation approaches?,'' in {\em Proceedings of the AAAI Conference on Artificial Intelligence}, vol.~35, pp.~11079--11087, 2021.

\bibitem{singh2021improving}
A.~Singh, N.~Doraiswamy, S.~Takamuku, M.~Bhalerao, T.~Dutta, S.~Biswas, A.~Chepuri, B.~Vengatesan, and N.~Natori, ``Improving semi-supervised domain adaptation using effective target selection and semantics,'' in {\em Proceedings of the IEEE/CVF Conference on Computer Vision and Pattern Recognition}, pp.~2709--2718, 2021.

\bibitem{yang2021deep}
L.~Yang, Y.~Wang, M.~Gao, A.~Shrivastava, K.~Q. Weinberger, W.-L. Chao, and S.-N. Lim, ``Deep co-training with task decomposition for semi-supervised domain adaptation,'' in {\em Proceedings of the IEEE/CVF International Conference on Computer Vision}, pp.~8906--8916, 2021.

\bibitem{qin2021contradictory}
C.~Qin, L.~Wang, Q.~Ma, Y.~Yin, H.~Wang, and Y.~Fu, ``Contradictory structure learning for semi-supervised domain adaptation,'' in {\em Proceedings of the 2021 SIAM International Conference on Data Mining (SDM)}, pp.~576--584, SIAM, 2021.

\bibitem{jiang2020bidirectional}
P.~Jiang, A.~Wu, Y.~Han, Y.~Shao, M.~Qi, and B.~Li, ``Bidirectional adversarial training for semi-supervised domain adaptation.,'' in {\em IJCAI}, pp.~934--940, 2020.

\bibitem{ma2022context}
N.~Ma, J.~Bu, L.~Lu, J.~Wen, S.~Zhou, Z.~Zhang, J.~Gu, H.~Li, and X.~Yan, ``Context-guided entropy minimization for semi-supervised domain adaptation,'' {\em Neural Networks}, vol.~154, pp.~270--282, 2022.

\bibitem{oord2018representation}
A.~v.~d. Oord, Y.~Li, and O.~Vinyals, ``Representation learning with contrastive predictive coding,'' {\em arXiv preprint arXiv:1807.03748}, 2018.

\bibitem{grill2020bootstrap}
J.-B. Grill, F.~Strub, F.~Altch{\'e}, C.~Tallec, P.~Richemond, E.~Buchatskaya, C.~Doersch, B.~Avila~Pires, Z.~Guo, M.~Gheshlaghi~Azar, {\em et~al.}, ``Bootstrap your own latent-a new approach to self-supervised learning,'' {\em Advances in neural information processing systems}, vol.~33, pp.~21271--21284, 2020.

\bibitem{khosla2020supervised}
P.~Khosla, P.~Teterwak, C.~Wang, A.~Sarna, Y.~Tian, P.~Isola, A.~Maschinot, C.~Liu, and D.~Krishnan, ``Supervised contrastive learning,'' {\em Advances in neural information processing systems}, vol.~33, pp.~18661--18673, 2020.

\bibitem{chen2020simple}
T.~Chen, S.~Kornblith, M.~Norouzi, and G.~Hinton, ``A simple framework for contrastive learning of visual representations,'' in {\em International conference on machine learning}, pp.~1597--1607, PMLR, 2020.

\bibitem{li2021probabilistic}
J.~Li, Y.~Zhang, Z.~Wang, and K.~Tu, ``Probabilistic contrastive learning for domain adaptation,'' {\em arXiv preprint arXiv:2111.06021}, 2021.

\bibitem{huo2021heterogeneous}
X.~Huo, L.~Xie, L.~Wei, X.~Zhang, X.~Chen, H.~Li, Z.~Yang, W.~Zhou, H.~Li, and Q.~Tian, ``Heterogeneous contrastive learning: Encoding spatial information for compact visual representations,'' {\em IEEE Transactions on Multimedia}, vol.~24, pp.~4224--4235, 2021.

\bibitem{zhang2022semi}
Y.~Zhang, X.~Zhang, J.~Li, R.~Qiu, H.~Xu, and Q.~Tian, ``Semi-supervised contrastive learning with similarity co-calibration,'' {\em IEEE Transactions on Multimedia}, 2022.

\bibitem{huang2024learningdifferentsamplessourcefree}
X.~Huang, C.~Zhu, B.~Zhang, and S.~Zhang, ``Learning from different samples: A source-free framework for semi-supervised domain adaptation,'' 2024.

\bibitem{yang2022class}
F.~Yang, K.~Wu, S.~Zhang, G.~Jiang, Y.~Liu, F.~Zheng, W.~Zhang, C.~Wang, and L.~Zeng, ``Class-aware contrastive semi-supervised learning,'' in {\em Proceedings of the IEEE/CVF Conference on Computer Vision and Pattern Recognition}, pp.~14421--14430, 2022.

\bibitem{he2020momentum}
K.~He, H.~Fan, Y.~Wu, S.~Xie, and R.~Girshick, ``Momentum contrast for unsupervised visual representation learning,'' in {\em Proceedings of the IEEE/CVF conference on computer vision and pattern recognition}, pp.~9729--9738, 2020.

\bibitem{berthelot2019mixmatch}
D.~Berthelot, N.~Carlini, I.~Goodfellow, N.~Papernot, A.~Oliver, and C.~A. Raffel, ``Mixmatch: A holistic approach to semi-supervised learning,'' {\em Advances in neural information processing systems}, vol.~32, 2019.

\bibitem{zhang2020does}
L.~Zhang, Z.~Deng, K.~Kawaguchi, A.~Ghorbani, and J.~Zou, ``How does mixup help with robustness and generalization?,'' {\em arXiv preprint arXiv:2010.04819}, 2020.

\bibitem{carratino2022mixup}
L.~Carratino, M.~Ciss{\'e}, R.~Jenatton, and J.-P. Vert, ``On mixup regularization,'' {\em The Journal of Machine Learning Research}, vol.~23, no.~1, pp.~14632--14662, 2022.

\bibitem{papyan2020prevalence}
V.~Papyan, X.~Han, and D.~L. Donoho, ``Prevalence of neural collapse during the terminal phase of deep learning training,'' {\em Proceedings of the National Academy of Sciences}, vol.~117, no.~40, pp.~24652--24663, 2020.

\bibitem{ding2023proxymix}
Y.~Ding, L.~Sheng, J.~Liang, A.~Zheng, and R.~He, ``Proxymix: Proxy-based mixup training with label refinery for source-free domain adaptation,'' {\em Neural Networks}, vol.~167, pp.~92--103, 2023.

\bibitem{bai2021understanding}
Y.~Bai, E.~Yang, B.~Han, Y.~Yang, J.~Li, Y.~Mao, G.~Niu, and T.~Liu, ``Understanding and improving early stopping for learning with noisy labels,'' {\em Advances in Neural Information Processing Systems}, vol.~34, pp.~24392--24403, 2021.

\bibitem{song2019prestopping}
H.~Song, M.~Kim, D.~Park, and J.-G. Lee, ``Prestopping: How does early stopping help generalization against label noise?,'' 2019.

\bibitem{grandvalet2004semi}
Y.~Grandvalet and Y.~Bengio, ``Semi-supervised learning by entropy minimization,'' {\em Advances in neural information processing systems}, vol.~17, 2004.

\bibitem{singh2021clda}
A.~Singh, ``Clda: Contrastive learning for semi-supervised domain adaptation,'' in {\em Advances in Neural Information Processing Systems} (M.~Ranzato, A.~Beygelzimer, Y.~Dauphin, P.~Liang, and J.~W. Vaughan, eds.), vol.~34, pp.~5089--5101, Curran Associates, Inc., 2021.

\bibitem{peng2019moment}
X.~Peng, Q.~Bai, X.~Xia, Z.~Huang, K.~Saenko, and B.~Wang, ``Moment matching for multi-source domain adaptation,'' in {\em Proceedings of the IEEE International Conference on Computer Vision}, pp.~1406--1415, 2019.

\bibitem{simonyan2014very}
K.~Simonyan and A.~Zisserman, ``Very deep convolutional networks for large-scale image recognition,'' {\em arXiv preprint arXiv:1409.1556}, 2014.

\bibitem{cubuk2020randaugment}
E.~D. Cubuk, B.~Zoph, J.~Shlens, and Q.~V. Le, ``Randaugment: Practical automated data augmentation with a reduced search space,'' in {\em Proceedings of the IEEE/CVF conference on computer vision and pattern recognition workshops}, pp.~702--703, 2020.

\bibitem{paszke2019pytorch}
A.~Paszke, S.~Gross, F.~Massa, A.~Lerer, J.~Bradbury, G.~Chanan, T.~Killeen, Z.~Lin, N.~Gimelshein, L.~Antiga, {\em et~al.}, ``Pytorch: An imperative style, high-performance deep learning library,'' {\em Advances in neural information processing systems}, vol.~32, 2019.

\bibitem{van2008visualizing}
L.~Van~der Maaten and G.~Hinton, ``Visualizing data using t-sne.,'' {\em Journal of machine learning research}, vol.~9, no.~11, 2008.

\bibitem{selvaraju2017grad}
R.~R. Selvaraju, M.~Cogswell, A.~Das, R.~Vedantam, D.~Parikh, and D.~Batra, ``Grad-cam: Visual explanations from deep networks via gradient-based localization,'' in {\em Proceedings of the IEEE international conference on computer vision}, pp.~618--626, 2017.

\bibitem{chen2019transferability}
X.~Chen, S.~Wang, M.~Long, and J.~Wang, ``Transferability vs. discriminability: Batch spectral penalization for adversarial domain adaptation,'' in {\em International conference on machine learning}, pp.~1081--1090, PMLR, 2019.

\bibitem{xie2022collaborative}
B.~Xie, S.~Li, F.~Lv, C.~H. Liu, G.~Wang, and D.~Wu, ``A collaborative alignment framework of transferable knowledge extraction for unsupervised domain adaptation,'' {\em IEEE Transactions on Knowledge and Data Engineering}, 2022.

\end{thebibliography}

\end{document}